%% file: arxiv-main.tex
\documentclass{article}

\usepackage{amsmath,amsfonts}
\usepackage{amsthm,amssymb,amsopn}
\newcommand{\expectation}{\ensuremath{\mathbb{E}}}
\newcommand{\Expt}{\expectation}

\usepackage{arxiv}

\usepackage[utf8]{inputenc} % allow utf-8 input
\usepackage[T1]{fontenc}    % use 8-bit T1 fonts
\usepackage[pdftex,bookmarks=true,pdfstartview=FitH,colorlinks,linkcolor=blue,filecolor=blue,citecolor=blue,urlcolor=blue,pagebackref=false]{hyperref}
\usepackage{url}            % simple URL typesetting
\usepackage{booktabs}       % professional-quality tables
\usepackage{amsfonts}       % blackboard math symbols
\usepackage{nicefrac}       % compact symbols for 1/2, etc.
\usepackage{microtype}      % microtypography
\usepackage{xcolor}         % colors
\usepackage{multirow}
\usepackage{graphicx}
\usepackage{subcaption}
\usepackage{mathnotation}
\usepackage{wrapfig}
\usepackage{enumitem}
\input{helpers}

\title{\systemName: Multi-level Ensemble Learning for \\ 
Resource-Constrained Environments}

\author{
Krishna Praneet Gudipaty$^1$ \\ %University of Massachusetts\\ Amherst, USA
\and Walid A. Hanafy$^1$ %\\ University of Massachusetts \\ Amherst, USA
\and Kaan Ozkara$^2$ %\\ University of California \\ Los Angeles, USA
\and Qianlin Liang$^4$\thanks{Work done while Qianlin Liang was a Ph.D. student at the University of Massachusetts Amherst.}
\and Jesse Milzman$^3$ %\\ Army Research Laboratory \\ USA
\and Prashant Shenoy$^1$ %\\ University of Massachusetts \\ Amherst, USA
\and Suhas Diggavi$^2$ %\\ University of California \\ Los Angeles, USA
}

\date{$^1$University of Massachusetts Amherst, USA\\
$^2$ University of California, Los Angeles, USA\\
$^3$ Army Research Laboratory, USA\\
$^4$ Amazon, USA}

\begin{document}

\maketitle

\begin{abstract}
  \input{sections/0-abstract}
\end{abstract}

%\section{Introduction - [UMass + UCLA] (1.75 Pages)}
\section{Introduction }
\label{sec:intro}

\input{sections/1-introduction}

%\section{Problem Formulation - [UCLA] (1.5 Pages)}
\section{Problem Formulation}
\label{sec:problem}
\input{sections/2-problem}

\section{System Architecture}% - [UMass + UCLA] (2 Pages)}
\label{sec:system}
\input{sections/3-system}

\section{Evaluation}% - [UMass] - (3.5 Pages)}
\label{sec:eval}
\input{sections/4-evaluation}

%\clearpage
\section{Conclusion}% - (0.25 Pages)}
\label{sec:conclusion}

\input{sections/5-conclusion}

% \begin{ack}

% \end{ack}
\bibliographystyle{plain}
\bibliography{main, Ref_DNNbibs}

\newpage
\appendix
\section*{Appendix}
\input{sections/appendix}
\end{document}

%% file: helpers.tex
\usepackage{titlesec}
\usepackage{wrapfig}
\usepackage{subcaption}
\usepackage{lipsum}

\titleformat{\subsubsection}[runin]
{\normalfont\bfseries}{\thesubsection}{1em}{}

\newcounter{todonumber}
\stepcounter{todonumber}

\def\systemName{\texttt{MEL}\xspace}

%% file: sections/0-abstract.tex
AI inference at the edge is becoming increasingly common for low-latency services. However, edge environments are power- and resource-constrained, and susceptible to failures. Conventional failure resilience approaches, such as cloud failover or compressed backups, often compromise latency or accuracy, limiting their effectiveness for critical edge inference services.  In this paper, we propose Multi-Level Ensemble Learning (\systemName), a new framework for resilient edge inference that simultaneously trains multiple lightweight backup models
capable of operating collaboratively, refining each other when multiple servers are available, and independently under failures while maintaining good accuracy. Specifically, we formulate our approach as a multi-objective optimization problem with a loss formulation that inherently encourages diversity among individual models to promote mutually refining representations, while ensuring each model maintains good standalone performance. Empirical evaluations across vision,language, and audio datasets show that \systemName provides performance comparable to original architectures while also providing fault tolerance and deployment flexibility across edge platforms.  Our results show that our ensemble model, sized at 40\% of the original model, achieves similar performance, while preserving 95.6\% of ensemble accuracy in the case of failures when trained using \systemName.

%% file: sections/1-introduction.tex
Edge AI has recently emerged as a complementary approach to cloud-based AI for running low-latency inference services, particularly due to the reduced network latency to access edge services from client devices~\cite{Satya2021:TheRole,Li2020:EdgeAI, Liang2023:Queueing,
Satya17_emergence}.  However, edge servers are often deployed as small modular data centers in harsh environments, such as the base of cell towers, making them more susceptible to hardware, connectivity, or power failures. Consequently, failures of edge AI services running on these platforms are likely to be more commonplace or more frequent than corresponding cloud-based services, necessitating the use of failure resilience 
techniques to withstand faults ~\cite{Soifer2019:MSFTInference, koren2007faultbook}. However, conventional failure resilience techniques that are commonly used in cloud platforms do not work well in edge settings. For example, the conventional approach of using backup replica servers to take over inference tasks from primary servers that experience failures may  be infeasible in edge data centers, since additional servers may be unavailable in such resource-constrained environments. 
Another approach is to use cloud resources, where the failure of one or more edge servers involves switching over to cloud servers~\cite{Hanafy2023:Degradation}, but doing so negates the latency advantage of edge inference. 
Yet another approach involves running a compressed version of the primary model on the limited resources available on a backup edge server and switching to the compressed model upon failure~\cite{wu2025:FailLite}. However, the size of a \emph{single} compressed backup model may be severely limited in resource-constrained environments, and doing so can degrade inference accuracy upon a failure.  
Thus, conventional approaches for handling edge failures come with several limitations. 

To address the issue of failover in resource-constrained environments, we propose a novel \emph{multi-level ensemble learning} methodology that can gracefully degrade performance with increasing failures. Our solution is motivated by the fact that while resources may be too constrained to run a single performant backup model, combining resources from different servers (\emph{e.g.,} by piggybacking on multiple servers) can allow us to achieve reasonable performance.
In particular, the goal of multi-level ensemble learning (\systemName) is to design models that: 1) refine each other so as to get better performance when combined (\emph{i.e.,} when more resources are available) and 2) are individually good to make the system more robust in case of failures.
In this paper, we investigate the approach of using multiple (i.e., two or more) such small failover backup models, which refine each other when combined. For example, we can use two small backup models $h_{\{1\}},h_{\{2\}}$, which are placed on two different edge servers $1$ and $2$, and are invoked when the main model $h_{\mathrm{\texttt{main}}}$ fails. Now, if the main server fails and only one of the two edge servers work, then we want an inference with just one of the small models; therefore for the backup models want ``reasonable'' performance individually. One trivial solution is to duplicate the small backup model in edge servers $1$ and $2$, and design it to get ``reasonable'' performance; however, in the event that both of the backup servers are available, then we do not gain any advantage. Therefore, the paper is focused on addressing the question of ensuring that the ensemble models are \emph{diverse} so that  they provide better performance when combined, while also providing reasonable performance individually.

Note that this is different from the popular mixture-of-experts paradigm \cite{shazeer2017}, where the individual learners are experts in certain aspects, and can perform very poorly in others. In contrast, we need the individual models to be reasonable learners in all aspects. Therefore, the first challenge to address is the loss criterion to be optimized to ensure such a design, for example, how to encourage diversity in the individual models; we will explore this in Section \ref{sec:problem}. The second challenge is to design the architecture and characteristics of the ensemble to be used for training to achieve the above properties; we will explore this in Section \ref{sec:system}.

\noindent\textbf{Contributions.} Our contributions can be summarized as follows:
\begin{itemize}[leftmargin=*]
    \item We formalize \systemName, a framework that jointly trains multiple lightweight backup models via a weighted combined training objective that also supports hierarchical labeling. \systemName is deployed with a fail-aware inference protocol that enables graceful performance degradation. Furthermore, we provide an information-theoretic generalization bound linking model diversity to complexity.  
    \item  
    We design the architecture of our ensemble models by using prefix blocks of the primary model architecture and detail the design space for selecting a suitable ensemble structure, as well as the runtime considerations. 
    \item We extensively evaluate \systemName across vision and audio datasets, showing our design flexibility across different resource constraints.  Our results show that our ensemble model, sized at 40\% of the original model, achieves similar performance, while preserving 95.6\% of ensemble accuracy in the case of failures.
    \item Lastly, we implemented a failure-resilient inference service based on our proposed architecture. Our evaluation in real edge settings showed that our \systemName approach not only achieves higher resilience than baseline approaches but also achieves 25\% less inference time.
\end{itemize}

\noindent\textbf{Related work.}
In \systemName, we propose a different approach that trains an ensemble of models capable of running cooperatively and independently, providing higher resilience guarantees. The prior work related to these aspects is addressed in this section. 

\textbf{AI Inference Systems.} 
AI inference systems allow users to deploy their models in the cloud, edge, and on-device platforms. Tools such as Nvidia Triton~\cite{nvidia_triton}, Kubeflow~\cite{kubeflow2025}, TensorFlow Serving~\cite{Olston2017:Tensorflow_serving}, TVM~\cite{Chen2018:TVMCompiler}, and ONNX Runtime~\cite{onnxruntime} allow users to deploy DNN models on a cluster of multiple cloud or edge resources. Building on these tools, researchers have focused on optimizing the inference latency and cost~\cite{Daniel2017:Clipper, Gujarati2020:Clockwork, Soifer2019:MSFTInference, Zhang2019_HeteroEdge, Samplawski2020:Towards, Zhang2021:DeepSlicing, Liang2023:Queueing, Gujarati2020:Clockwork, Zhang202:MArk, Ahmad2024:Loki, 
Hanafy2023:GPUDirect,
Liang2020:AIEge} and addressing challenges of workload dynamics~\cite {Ahmad2024:Proteus, Ahmad2024:Loki, Zhang2020:ModelSwitching, Wan2020:ALERT, Liang2023:Delen}. \systemName builds on these efforts and addresses the issue of failure resiliency that has not seen much attention, especially in resource-constrained environments.

\textbf{Failure Resiliency in Resource-constrained Environments.} 
In cloud platforms, resources are abundant, and traditional failure resiliency approaches, such as replication, can be employed. For instance, in \cite{Soifer2019:MSFTInference}, the author replicates every request to optimize the latency and ensure failure resiliency. 
In contrast, edge environments, which are more prone to power and hardware failures than cloud resources, as they are typically deployed in remote and sometimes hostile environments, are highly resource-constrained, rendering such replication approaches inapplicable.  
A widely accepted strategy for ensuring failure resilience in resource-constrained environments is graceful degradation~\cite{graceful_2023, defcon,koren2007faultbook, Hanafy2023:Degradation}, which often involves sacrificing performance by limiting the available functionalities. For example,~\cite{defcon, graceful_2023} allocates resources according to the criticality of applications, reducing the number of services and functions available during failures. In the context of AI inference, researchers have focused on two main techniques. The first approach involves model compression and selection~\cite{wu2025:FailLite, Hanafy2023:Degradation}, where models are substituted with smaller backups that usually use fewer resources but offer lower accuracy. In this case, when the primary (\emph{i.e.,} large) model fails, the inference requests are redirected to a small failover replica. 
The second approach utilizes early-exit and skip-connection to horizontally split models processing across servers to guarantee a result is produced even when one or more servers fail~\cite{yousefpour2019guardians, yousefpour2020resilinet, majeed2022continuer}. 
In contrast to these approaches, \systemName trains an ensemble model capable of harnessing resources across multiple servers. \systemName provides a complementary approach, where users can select the ensemble size and allow for the sub-models to be placed on different servers. 

\textbf{Ensemble Learning.} 
Early ensemble methods such as Bagging \cite{breiman1996bagging}, AdaBoost \cite{freund1997decision} and Random Forests \cite{breiman2001random} established that aggregating diverse base learners reduces variance and boosts accuracy by voting or averaging predictions. These approaches inspired a large body of work on model aggregation, stacking that remains a cornerstone of statistical learning theory. However, they do not consider resilience or robustness towards failures of individual learners.  Orthogonally, Mixture-of-Experts (MoE) architectures route each input through a subset of specialist subnetworks, achieving conditional computation and massive parameter scaling. MoE dates back to \cite{jordan1994hierarchical}, recently \cite{shazeer2017} transformed MoE from a modest modular learner into a scalable conditional‑compute layer by introducing sparse top‑k routing with load‑balancing regularizers and an efficient parallel implementation, enabling trillion‑parameter‑level capacity at roughly the cost of a dense baseline. MoE, as the name suggests, is made up of `experts' that specializes according to inputs. In contrast, in \systemName, we require each learner to be generalizable to inputs so that the failure of a learner is not catastrophic (as it would be in MoE).

\begin{figure}[t]
    \centering
    \includegraphics[width=0.7\linewidth]{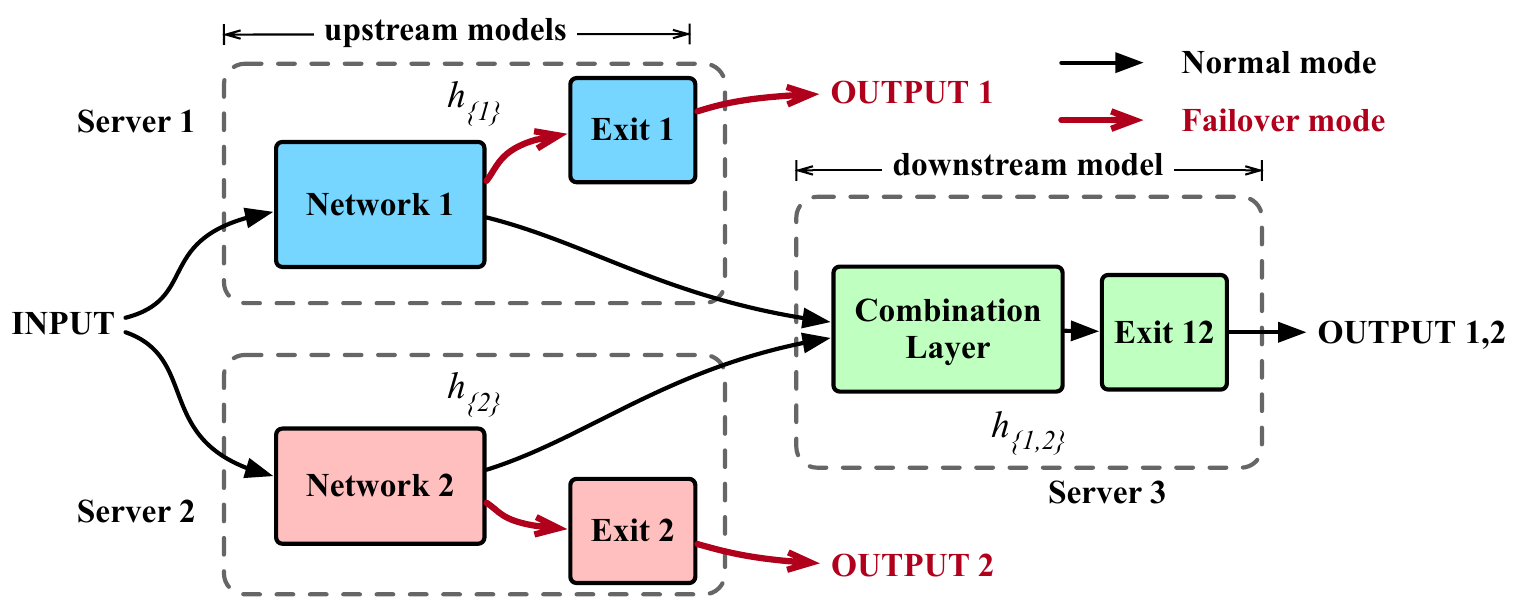}
    \caption{An example \emph{multi-level ensemble} architecture featuring two upstream models and a downstream model comprising a combination and exit layers. The figure also highlights the data flow in the normal mode, where servers 1,2, and 3 are available, and in the failover mode, where either one of the servers or the combination server (i.e., server 3) fails.}
    \label{fig:ensemble_arch}
\end{figure}

%% file: sections/2-problem.tex
To formalize the \systemName problem, 
%To set up  some standard notation,
let $\vct{x}$ be inputs to the learning model and $y$ be the labels associated with it in the supervised learning framework. We will consider learning models under a loss function $\ell(h(\vct{x}),y)$, where $h(\vct{x})$ is the prediction by model $h$ for input $\vct{x}$. Conventionally, the performance is measured by the population risk as $L(h)=\Expt_{(\vct{x},y)\sim\mathcal{P}}[\ell(h(\vct{x}),y)]$, for an unknown underlying distribution $\mathcal{P}$, and the model $h$ is optimized over a hypothesis class $h\in\mathcal{H}$.

In our problem we are simultaneously training multiple models $\{h_{\mathcal{S}}\}$, where $\mathcal{S}\subseteq \mathcal{M}=\{1,\ldots,M\}$, where we would like to ensure that the performance of each of the models is maintained while refining each other. That is, we would like to design $\{h_{\mathcal{S}}\}$ such that the population risks $L(h_{\mathcal{S}})\leq \gamma_{\mathcal{S}}$, where we also want $\gamma_{\mathcal{S}}\leq \gamma_{\mathcal{S}'}$ if $\mathcal{S}'\subset \mathcal{S}$, \emph{i.e.,} we have refinement in performance when we have access to more backup servers. To make this concrete, from \autoref{fig:ensemble_arch}, we have two backup servers, and the individual models are $h_{\{1\}},h_{\{2\}}$, and when both backup models succeed, they feed into another server which can build a model $h_{\{1,2\}}$, which takes intermediate representations from $h_{\{1\}},h_{\{2\}}$ and does further processing to produce the final inference. The goal is to simultaneously train $h_{\{1\}},h_{\{2\}},h_{\{1,2\}}$ for the desired individual performances. We will use the terminology upstream model for the models that feed into later downtream models; for \emph{e.g.,} in this case $h_{\{1\}},h_{\{2\}}$ are upstream models feeding into the downstream model $h_{\{1,2\}}$.

\textbf{Objective:} Note that we also want the model classes to be smaller (or less complex) for smaller subsets, \emph{i.e.,} $h_{\mathcal{S}}\in \mathcal{H}_{\mathcal{S}}$ and the parametrization of $\mathcal{H}_{\mathcal{S}}$ is smaller if $\mathcal{S}$ is smaller. Clearly, this leads to several natural questions {\sf (i)} how do we compose the models $\{h_{\mathcal{S}}\}$, \emph{i.e.,} what is exchanged between the upstream models and the downstream models? {\sf (ii)} how do we choose the individual model classes $\mathcal{H}_{\mathcal{S}}$, \emph{i.e.,} architect the system {\sf (iii)} how do we train the models to satisfy these requirements? {\sf (iv)} can we say anything about generalization performance bounds for this problem setup?

\noindent\textbf{Training criterion:} We will address {\sf (i)} and {\sf (ii)} in Section \ref{sec:system}, but will set up the framework for {\sf (iii)} and {\sf (iv)} in this section next. Clearly to simultaneously find the models $\{h_{\mathcal{S}}\}$, it is a multi-objective optimization. For example we can setup
\begin{equation}
  \label{eq:LearningEnsemble}
  \min_{\{h_{\mathcal{S}}\in \mathcal{H}_{\mathcal{S}}\}} L(h_{\mathcal{M}})\,\, \mbox{such that } L(h_{\mathcal{S}})\leq \gamma_{\mathcal{S}}, \,\, \mathcal{S}\subseteq \mathcal{M}=\{1,\ldots,M\}
\end{equation}
As is conventional, we solve this by replacing the population risks by their empirical risks, and therefore we formulate this as a Lagrangian as follows:
\begin{equation}
  \label{eq:Emp-LearningEnsemble}
\mathcal{L} = \sum_{\mathcal{S}\subseteq \mathcal{M}} \lambda_{\mathcal{S}} \hat{L}(h_{\mathcal{S}}),
\end{equation}
where $\{\lambda_{\mathcal{S}}\}$ define the Lagrangian weights and we have defined the empirical risks over a labeled dataset $\mathcal{D}=\{(\vct{x}_1,y_1),\ldots,(\vct{x}_N,y_N)\}$ as
\begin{equation}
  \label{eq:Emp-Ensemble-Risks}
  \hat{L}(h_{\mathcal{S}}) = \frac{1}{N}\sum_{i=1}^N \ell(h_{\mathcal{S}}(\vct{x}_i^{(\mathcal{S})}),y_i^{(\mathcal{S})}).
\end{equation}
Note that we are allowing ourselves the flexibility to ``process'' the dataset $\mathcal{D}$ to produce datasets $\mathcal{D}^{(\mathcal{S})}=\{(\vct{x}_1^{(\mathcal{S})},y_1^{(\mathcal{S})}),\ldots,(\vct{x}_N^{(\mathcal{S})},y_N^{(\mathcal{S})})\}$. For example the processing we investigate in this paper is to ``coarsify'' the labels to produce $\{y_i^{(\mathcal{S})}\}$, \emph{i.e.,} can take the CIFAR-100 dataset and use their 20 coarser superclass labels. 
One could also compress the inputs $\{\vct{x}_i\}$ to produce the hierarchy specific inputs $\{\vct{x}_i^{(\mathcal{S})}\}$.
Therefore, our overall optimization problem to simultaneously find the models $\{h_{\mathcal{S}}\}$ from \eqref{eq:Emp-LearningEnsemble} becomes,
\begin{equation}
  \label{eq:Multilevel-Ensemble-Opt}
 \min_{\{h_{\mathcal{S}}\in \mathcal{H}_{\mathcal{S}}\}} \sum_{\mathcal{S}\subseteq \mathcal{M}} \lambda_{\mathcal{S}} \hat{L}(h_{\mathcal{S}}).
\end{equation}
The coefficients $\{\lambda_{\mathcal{S}}\}$ capture the relative importance of the multiple levels of this ensemble learning problem. Let us take a concrete example of $\mathcal{M}=\{1,2\}$, and the problem then becomes $\min_{h_{\{1\}},h_{\{2\}},h_{\{1,2\}}}\{\lambda_1 \hat{L}(h_{\{1\}})+\lambda_2 \hat{L}(h_{\{2\}})+\lambda_{12} \hat{L}(h_{\{1,2\}})\}$. If we increase $\lambda_{12}$ relative to $\lambda_1,\lambda_2$ then we want to encourage the individual models $h_{\{1\}},h_{\{2\}}$ to be ``different''/``diverse'' so that they can be mutually refined to produce better performance for the downstream model $h_{\{1,2\}}$.

\noindent\textbf{Inference time operation:} During training, we need to simultaneously design $\{h_{\mathcal{S}}\}$ to be prepared for any configuration of failures at inference time. However, at inference time we have access to knowledge of the failures and we activate the appropriate backup models. For example, if the main server allocated to the task fails, so does its model $h_{\mathrm{\texttt{main}}}$. Then we know which subset $\mathcal{S}$ of backup servers have succeeded, and hence know which model $h_{\mathcal{S}}$ can be used. At inference time, we can operate with the full precision input $\vct{x}$ and feed it to the model $h_{\mathcal{S}}$.

\noindent\textbf{Learning theoretic setup. } For simplicity we utilize the framework initiated in \cite{xu2017information} \footnote{Following this work, there are important works \cite{RammalAGDS22,harutyunyan21,steinke2020,bu2020information,russo16,haghifam2020} using conditional mutual information (CMI), which we do not focus on for simplicity. However, we believe CMI results can be translated to our work. }. We consider an instance space $\mathcal{Z}$, a dataset $\mathcal{D}=\{\vct Z_i\}_{i=1}^n$ where $(\vct{x}_i,y_i)=\vct Z_i\in \mathcal{Z}$ and each sample is sampled from an underlying distribution $\mathcal{P}$, i.e., $\vct Z_i\sim \mathcal{P}$, a hypothesis space $\mathcal{H}$, and a non-negative loss function $\ell:\mathcal{H}\times \mathcal{Z} \rightarrow \mathbb{R}^+$. A learning algorithm tries to output a hypothesis $h$ such that population loss $L(h):= \mathbb{E}_{Z\sim \mathcal{P}} [\ell(h,\vct Z)]$ is minimized. As $\mathcal{P}$ is unknown, empirically, we utilize the dataset $\mathcal{D}$ to minimize empirical risk, $\hat L(h):= \frac{1}{n} \sum_{i=1}^n \ell(h,\vct Z_i)$. Then, the generalization error can be characterized as $\text{gen}(\mathcal{P}, P_{h|\mathcal{D}}):= E_{h,\mathcal{D}}[L(h)-\hat L(h)]$, where $P_{h|\mathcal{D}}$ is the conditional distribution of output hypothesis given the input dataset.  

    \cite{xu2017information}, obtains a generalization bound that relates mutual information between the hypothesis and dataset to generalization performance. 

    \begin{lemma}[Theorem 1 in \cite{xu2017information}.]
        Given that $l(h,Z)$ is $\sigma-sub-Gaussian$ for all $h\in \mathcal{H}, Z\sim \mathcal{P}$ we have,
        \begin{align*}
            (\text{gen}(\mathcal{P}, P_{h|\mathcal{D}}))^2\leq \frac{2\sigma^2}{n}I(\mathcal{D};h).
        \end{align*}
    \end{lemma}
    In our setting we consider a Markov graph (see graph in Appendix) with 4 nodes in the form of $\mathcal{D}\rightarrow h_{\{1\}}, h_{\{2\}} \rightarrow h_{\{1,2\}}$, where individual hypotheses may be from different spaces $\mathcal{H}_1,\mathcal{H}_2,\mathcal{H}_{1,2}$; note given $\mathcal{D}$, $h_{\{1\}}$ and $h_{\{2\}}$ are independent. As a result can relate the generalization error with the mutual information $I(h_{\{1\}};h_{\{2\}})$ as in the next theorem,
    \begin{proposition} \label{thm}
         We consider overall generalization error $\text{gen}^2_{overall} := \mathbb{E}_{\mathcal{D},h_{\{1\}},h_{\{2\}},h_{\{1,2\}}}[L(h_{\{1,2\}},h_{\{1\}},h_{\{2\}})]=\mathbb{E}[\frac{1-p}{1+p}(L(h_{\{1,2\}})-\hat L (h_{\{1,2\}}))^2 + \frac{p}{1+p}(L(h_{\{1\}})-\hat L(h_{\{1\}}))^2+\frac{p}{1+p}(L(h_{\{2\}})-\hat L (h_{\{2\}})^2]$, where $0 \leq p\leq 1$ acts as a failover probability, i.e. when $p=1$ it is guaranteed to use a lightweight model, and $p=0$ it is guaranteed to use the ensemble model. Then the overall generalization error (sum of squares) yields the following upper bound,
        \begin{align*}
        \text{gen}^2_{overall} \leq \frac{1}{1+p}\frac{2\sigma^2}{n}((I(D;h_{\{1\}})+I(D;h_{\{2\}}))-(1-p)(I(h_{\{1\}};h_{\{2\}})))
    \end{align*}
    \end{proposition}
    
    \textbf{Remark. } The generalization error is tightly related to the model complexity/capability. A more capable model is also capable of resulting higher generalization error due to overfitting risk \cite{shalev2014}. In this remark, we tie the diversity of $h_{\{1\}},h_{\{2\}}$ to model complexity through the generalization error \cite{jakubovitz2019generalization}. We observe that the generalization error depends on $I(h_{\{1\}};h_{\{2\}})$, the overall complexity of the architecture can be characterized through the mutual information of $h_{\{1\}}$ and $h_{\{2\}}$. When $I(h_{\{1\}};h_{\{2\}})$ is small, i.e. $h_{\{1\}}$ and $h_{\{2\}}$ likely to output more diverse hypotheses, we observe that generalization error upper bound is increased indicating the increase in complexity as expected. And when $I(h_{\{1\}},h_{\{2\}})$ is high it means $h_{\{1\}}$ and $h_{\{2\}}$ highly related and convey `similar' information in the downstream channel, this indicates the overall architecture is less capable (complex) and the generalization error is decreased. Note that, given a capable model, overfitting can be mitigated through increasing/diversifying dataset. Furthermore, even though our method increases the generalization upper bound, we have \emph{population risk=empirical risk+generalization error}; hence, if empirical loss is decreased sufficiently due to increased capability, we can still observe decreased population risk.

%% file: sections/3-system.tex
In this section, we explain how we instantiate the ensemble model $h_{{\mathcal{S}}}$ (i.e., question {\sf (i)}) and explain the design space and run-time considerations (i.e., question {\sf (ii)}) from \autoref{sec:problem}.

\noindent \textbf{Ensemble Architecture.} Our multi-level ensemble $\{h_\mathcal{S}\}$ features mutiple upstream and downstream models as depicted in \autoref{fig:ensemble_arch}. We base the architecture of our upstream models on the concept of \emph{neural block}, or \emph{block} for short, a commonly used approach in designing modern neural networks~\cite{resnet, efficient-net}. 
Similar to multi-exit~\cite{Liang2023:Delen, Scardapane2020:WhyEarlyExit, BranchyNet} and split processing approaches~\cite{Huang2020:CLIO, Banitalebi2021:AutoSplit, Matsubara2022:SplitSurvery, Samplawski2020}, we design the upstream models by considering the original model and selecting a prefix of its blocks. For example, since the EfficientNet-B0 architecture comprises seven blocks, each with one or more convolution layers,  we can design the smaller upstream models as a subset of blocks using a prefix of the original model architecture, which limits our design space to a discrete set of options. Additionally, we attach an exit block (\emph{e.g.,} classification layer) to each upstream model that is trained as part of the ensemble and only activated in cases of failures.
Further, in contrast to early-exit or split processing, where subsequent phases or blocks rely on a single source of intermediate representation, our architecture combines intermediate representations from multiple complementary sources. Hence, our downstream model primarily contains a combination layer and output layer, but may feature a more complex architecture as we highlight in \autoref{app:ablation}.  
Lastly, as described in \autoref{sec:problem}, we train the ensemble by considering a multi-objective training criterion
that considers the accuracy of upstream models, their diversity, and the accuracy of the downstream model.

\noindent\textbf{Design Space and Considerations.}
Several design decisions govern the training of a multi-level ensemble model for a specific edge deployment scenario. We discuss the design space and the runtime considerations for these design decisions below, with additional details in \autoref{app:more_system_details}.

\begin{itemize}[leftmargin=*]
    \item Our \emph{first} design decision is the size and number of the upstream models, as both factors are key to their learning ability and resiliency. Since our approach uses the granularity of a block, only a discrete (and finite) set of choices is available to determine the number and sizes of upstream models. For example, the first block of a given architecture yields the smallest size upstream model $h_i$. In particular, our approach considers the available resources per server and selects a prefix subset of blocks based on this resource availability. As we demonstrate in \autoref{fig:size_effbet},  since the model size significantly impacts the accuracy and larger sizes come with diminishing returns,  picking a model size based on the knee of the curve yields a good balance between size and accuracy. 
    Additionally, increasing the number of upstream models increases training complexity but also enables the system to tolerate more failures.

    \item Our \emph{second} design decision is the size of the downstream model. 
    While system resource constraints play an important role in determining the model size, they are not the only factor influencing our design.
    Resource fragmentation can also play a critical role since the available resource budget may not be evenly split across servers. 
    In such scenarios, a key aspect of our design is how to apportion the total available resource between the upstream and downstream models.    
    On one hand, allocating more resources to the upstream model creates performant models that benefit less from being combined; on the other hand, allocating more resources to the downstream model creates models dependent on the availability of the combination model. 

    \item Our \emph{third} decision is the relative importance of the upstream and downstream models, determined by the coefficients $\lambda_{\mathcal{S}}$. 
    The relative importance determines the model's behavior in different scenarios. For instance, a higher weight for the downstream models encourages the ensemble model to produce ``different'' or ``diverse'' models that refine each other and increase its accuracy.
    In contrast, a higher weight for the upstream models trains more independent models that may not provide benefits when combined. 
    The relative importance is also essential for determining the system's resiliency under different failure types and the expected availability of different servers. 
    For instance, in scenarios where network connectivity is intermittent, larger (i.e., more independent) upstream models are more adequate as they are less affected by connectivity failures. 
    
    \item Our \emph{fourth} design decision is the output granularity, where users can opt for results that represent slightly different (\emph{i.e.,} less complex) problems.
    One particular aspect that we focus on is to ``coarsify'' the output labels, which allows the model to yield higher accuracy, even in severely constrained environments, as we show in \autoref{tab:hier}. 
    For example, users can choose only to consider a binary classifier rather than a multi-class classifier, or classify the image as an animal as opposed to whether it is a dog or a cat. 
\end{itemize}

In \autoref{sec:eval}, \autoref{app:more_models},\autoref{app:ablation}, we investigate the impact of the above design decisions and show the accuracy and the associated resource requirements of different design points. 

\noindent \textbf{\systemName Deployment.} 
In evaluating the performance of \systemName, we implement a failure-resilience inference service. Since, from a practical standpoint, resource availability and failures are not known a priori at training time, our approach trains a family of multi-level ensembles, each for a different structure and resource requirements. In this case, at runtime, users or servers employ failure detection mechanisms (e.g., via heartbeat and timeouts). If the primary model fails, the users can choose the most appropriate ensemble to deploy in a particular edge environment via a best-fit approach. We detail our system implementation in \autoref{app:more_system_details} and depict the performance of an exemplar deployment in \autoref{fig:respone_time}.

%% file: sections/4-evaluation.tex
% - What is the setup?.
% - Single modality/Single Architecture/ Strawman
% - Not very special and two-dimensional: arch, data sets, and modalities
% - Mutli-level
% - model-sizes
% - Effect of compression.
% - Abliation:
%     - Change the head
%     - Change the number of models

In this section, we extensively evaluate the performance of \systemName by comparing it across baselines. We start with an experimental setup, then introduce the evaluation results, followed by an ablation study of different configurations and scenarios.

\input{tables/datasets}
\subsection{Experimental Setup} \label{sec:eval_setup}
\subsubsection*{Datasets.} We evaluate our approach on the datasets listed in ~\autoref{tab:arxiv_datasets}, (1) CIFAR-100~\cite{krizhevsky2009:CIFAR}, consisting of 60K images across 100 fine-grained classes grouped into 20 superclasses, (2) Tiered ImageNet~\cite{Ren2018:Multi-Level}, a 700K-image subset of ImageNet-1k~\cite{Deng2009:imagenet} with 608 classes organized into 34 high-level categories, (3) Speech Commands~\cite{speechcommandsv2}, which contains 65K one-second audio recordings for keyword classification and speech recognition tasks, and (4) BookCorpus~/\cite{Zhu2015:Books}, which is comprised of 11K novels and converted to 1.2 billion tokens using a tokenizer with a vocab size of 8K.

\subsubsection*{DNN Architectures.} We utilize a multitude of DNN architectures to evaluate the performance of our multi-level ensemble learning methodology. In particular, we use EfficientNet-B0~\cite{efficient-net}, Resnet50~\cite{resnet}, ViT-B/16~\cite{vit}, and DeepSpeech2~\cite{amodei2015deepspeech2endtoend}. Our experiments denote these architectures as \emph{original}, and construct our multi-level ensemble models using sub-blocks of these models as explained in ~\autoref{sec:system}. %\kaan{better explanation of block notation is needed} 
To establish notation, we refer to ensemble models by their architecture and the number of blocks used, as our current evaluation only focuses on fully connected downstream models. For example, EfficientNet-B0 ($\Bc3$), denotes an ensemble architecture with upstream models that uses first three layers of EfficientNet-B0. %Since our current evaluation omits the effect of the downstream model, .  
% we will refer to the model that uses the first three sub-blocks of EfficientNet-B0 for the small backup model as EfficientNet-B0 ($\Bc3$). 

\subsubsection*{Baselines.}\label{sub:baselines} Our evaluation considers the following baselines: (1) the \textbf{original} architecture (for e.g. full-size Efficient-B0), which defines the target downstream model performance (2) a \textbf{small} failover replica, with the same size as $h_{\{1\}}$, serving as a baseline for conventional lightweight failover models~\cite{Hanafy2023:Degradation, wu2025:FailLite} (3) the \textbf{standalone} version of the ${h_{\mathcal{S}}}$ architecture, where only $h_{\{1,2\}}$ is optimized—serving as a benchmark for downstream accuracy when no constraints are imposed on upstream models (i.e., $\lambda_1 = \lambda_2 = 0$), and (4) an \textbf{individually-trained} ensemble, where the upstream models $h_{\{1\}}, h_{\{2\}}$ are trained independently in the first stage, followed by downstream training of $h_{\{1,2\}}$ with frozen weights in the upstream models, representing an alternative way to define $h_{\mathcal{S}}$ by combining representations post hoc, in contrast to the joint optimization used in \systemName. 
%\kaan{explain goal of these baselines, why these baselines, what is the purpose? }

\subsubsection*{Training Details.}
In addition to the joint optimization proposed by \systemName, we also fine-tune the downstream model while freezing upstream models to further improve accuracy, which will be discussed in the following sub-section.
%We train models in two steps: First, we train the ensemble as explained in \autoref{sec:problem} and then we fine-tune the downstream model while freezing the upstream model, which further enhances the accuracy of our model as we show in \autoref{tab:hier}.
All models are implemented using PyTorch v2.5.1 and optimized using AdamW~\cite{loshchilov2019:adamW} %\walid{AdamW} \kaan{give AdamW citation instead of Adam} 
under a cosine learning rate scheduler~\cite{loshchilov2017:sgdr}. Lastly, the assessment of the system performance and response time of DNN inference services based on these models is done using ONNX Runtime v1.21, gRPC v1.71 on a local cluster of three Nvidia A2 GPUs. 
Full details of our setup, models, and datasets are available in ~\autoref{app:exp_details}.
%\walid{We also should evaluate on CPUs} 

\subsection{Experimental Results}\label{sec:eval_main}
In this section, we present the experimental results of \systemName, focusing on ensembles composed of \emph{two} upstream models.
%In this section, we present the main results of \systemName while focusing on two model ensembles. 
\autoref{app:more_models} presents the full results of \systemName across architectures and datasets.% are presented in \autoref{app:more_models}. %\autoref{app:ablation}, and \autoref{app:baselines}.
% \kaan{1. very clean first result that demonstrates our main motivation, on a single dataset}\\
% \kaan{2. ablations/datasets that indicates that our result generalizes across the board}\\
% \kaan{3. comparison to baselines/strawmen}\\
% \kaan{4. system level results?}\\

\subsubsection*{Initial Evaluation.}

% \begin{wrapfigure}{R}{0.4\textwidth}
%   \centering
%   \includegraphics[width=0.38\textwidth]{images/cifar100_effb0_5.pdf}
%   \caption{Comparing accuracy across different baselines (model size on top) \kaan{I thought we were gonna change this?}.}
%   \label{fig:result_effb0_5}
% \end{wrapfigure}

\input{tables/results-intro}

\autoref{tab:intro_results} illustrates the results of our training approach using CIFAR-100 with EfficientNet-B0 ($\Bc5$), i.e., using the first five blocks, and Resnet50 ($\Bc3$). The table shows the Top-1 accuracy and the number of parameters.\footnote{For $h_{\{1,2\}}$, the number of parameters represents the whole ensemble including exits for $h_{\{1\}}$ and $h_{\{2\}}$.}
As shown, the results demonstrate that our proposed method is comparable to the accuracy of the original model, where $h_{\{1\}}$ and $h_{\{2\}}$ refine one another, leading to improved final downstream accuracy for $h_{\{1,2\}}$. 
Using our approach, the individual upstream models' accuracy is not sacrificed, ensuring that the system remains robust—if server 1 or server 2 fail, either will perform reasonably well as a failover mechanism. For instance, the table shows that upstream models can achieve up to 95.6\% of ensemble accuracy in case of failures.
Also, we are able to achieve this effectively with a reasonably smaller number of parameters (translates to lesser memory and computational requirements),  providing an additional benefit of using this strategy. 
% More importantly, the figure emphasizes that our approach does not sacrifice the accuracy of the upstream models ($h_{\{1\}}$ and $h_{\{2\}}$)  \kaan{emphasize resilience more}. Thus, if server 1 or server 2 fails, either will perform reasonably well as a standalone model.

\input{tables/models-datasets}
\subsubsection*{Overall Performance.}
Results across a broader evaluation, as in \autoref{tab:arxiv_generalization}, show that our training approach is effective in creating performant upstream and downstream models across datasets and model architectures. The combined downstream model $h_{\{1,2\}}$ behaves similarly and sometimes \emph{better} than the original model with a smaller parameter budget. For instance, Efficient-B0 ($\Bc3$) trained on the Mel-Spectrogram of Speech Commands dataset behaves similarly to the original model, despite only having around \emph{15}\% of the parameters. In cases where we achieve the performance with a similar range of parameters when compared to the original, there is design flexibility, in the sense that the ensemble model trained using \systemName approach can better utilize the same parameter budget (i.e., memory requirements), with a more resilient execution towards failures.
%For instance, Efficient-B0($\Bc6$) trained on TieredImageNet has a similar size to the original, but its ensemble nature makes it more resilient to failures.  
%\kaan{need comment about gpt}

\begin{figure}[t]
  \centering
  \hfill
  \begin{subfigure}[b]{0.45\textwidth}
    \centering
    \includegraphics[width=0.8\textwidth]{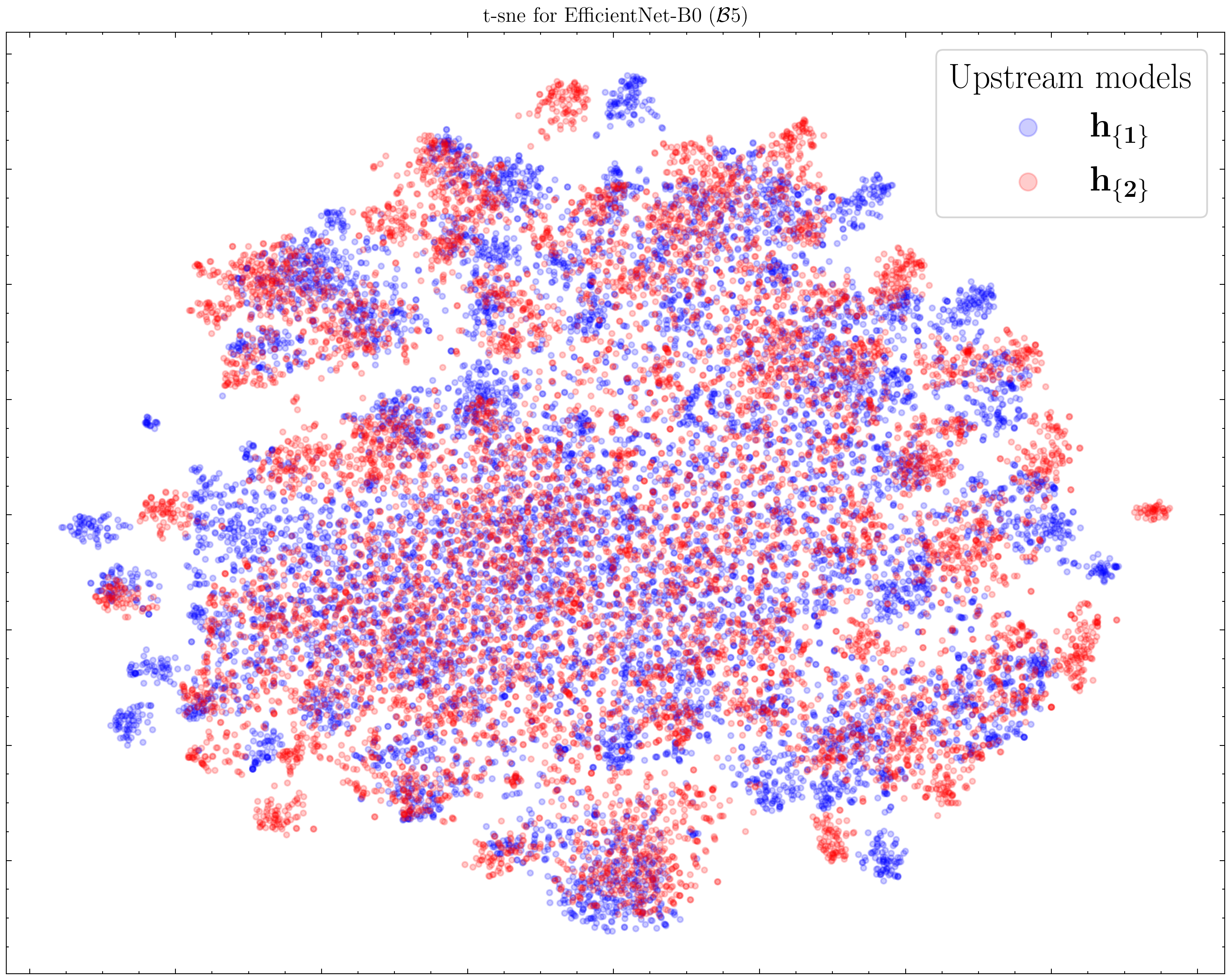}
    \caption{CIFAR-100}
    \label{fig:tsne_effnet_c5_cifar}
  \end{subfigure}
  \hfill
  \begin{subfigure}[b]{0.45\textwidth}
    \centering
    \includegraphics[width=0.8\textwidth]{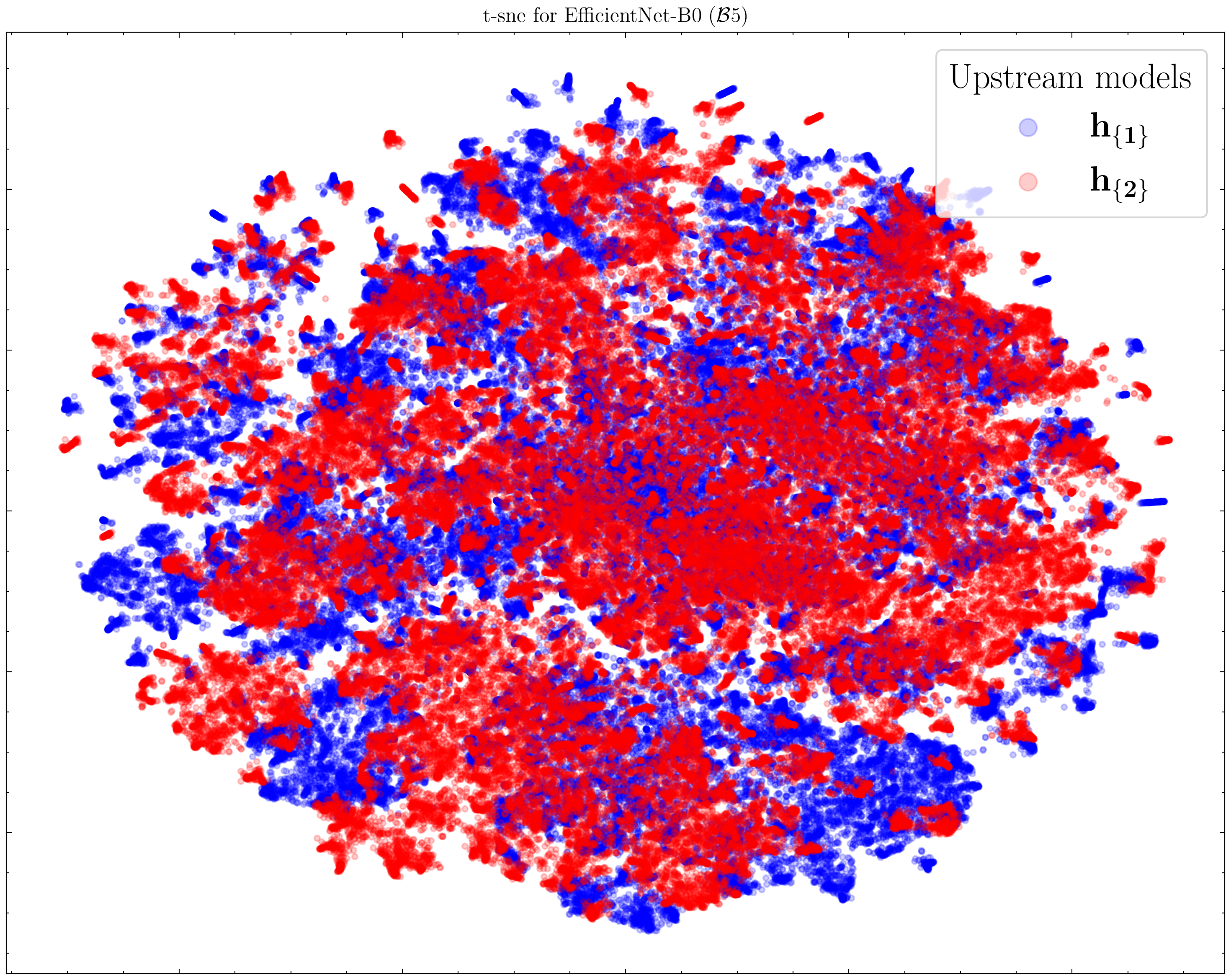}
    \caption{TieredImageNet}
    \label{fig:tsne_effnet_c5_tin}
  \end{subfigure}
  \hfill
  \hfill
    \caption{t-SNE visualization of features from upstream models $h_{\{1\}}$, $h_{\{2\}}$ in the \systemName framework, distinguished by color, for EfficientNet-B0 ($\Bc5$) on the CIFAR-100 and TieredImageNet datasets.}
    \label{fig:tsne_visuation}
\end{figure}

\subsubsection*{Feature visualization} 
To qualitatively assess the behavior of models trained under the \systemName framework, we visualize the learned feature embeddings from upstream models $h_{{1}}$ and $h_{{2}}$ using t-SNE dimensionality reduction on CIFAR-100 and TieredImageNet datasets as shown in ~\autoref{fig:tsne_visuation}. As expected, there is and overlap that exists between the representations, but also the embeddings reveal that the models also capture complementary information. This indicates that during \systemName training, the upstream models learn some mutually refining information/diversity among upstream models, enabling them to enhance downstream performance when combined by $h_{\{1,2\}}$ by covering different aspects of the input space.

\subsection{Ablation studies}\label{sec:eval_ablation}
%\walid{We only have one item in ablation, maybe move this part to the earlier section.}
% \begin{wrapfigure}{r}{0.32\linewidth}
% \centering
%     %\begin{subfigure}[t]{\linewidth}
%        % \centering
%     % \vspace{-1.6cm}
%     \includegraphics[width=\linewidth]{images/effnet_cifar_ensemble_accuracy_vs_parameters.pdf}
%         \caption{Effect of ensemble size of EfficientNet-B0 on CIFAR-100.}
%         \label{fig:size_effbet}
%         \vspace{-0.5cm}
%     %\end{subfigure}
%     % \newline
%     % \begin{subfigure}[t]{\linewidth}
%     %     \centering
%     % \includegraphics[width=\textwidth]{images/cifar100_effb0_5_weights.pdf}
%     %     \caption{Increasing weight on ensemble performance during training}
%     %     \label{fig:weights}
%     % \end{subfigure}
%     % \caption{Main figure caption describing both subfigures}
%     %\todo{Comment on the architecture and dataset in the figure/text.}
%     % \walid{We should pick 1 from (a) and (b). The figure (c) is not what we expect. For some reason, adding more weight makes $h_{12}$ worse and also 1 of them. I think the reason is that it tries to make it as a regular model.}
%     \label{fig:main}
% \end{wrapfigure}
\subsubsection*{Effect of model size.} 
\begin{figure}
    \centering
    \hfill
    \begin{minipage}[b]{0.45\textwidth}
    \centering
     \includegraphics[width=0.8\linewidth]{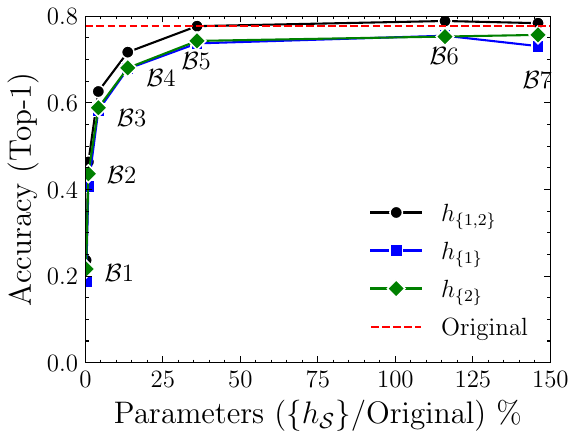}
        \caption{Effect of ensemble size of EfficientNet-B0 on CIFAR-100.}
        \label{fig:size_effbet}
    \end{minipage}
    \hfill
      \begin{minipage}[b]{0.45\textwidth}
    \centering
    \includegraphics[width=0.88\linewidth]{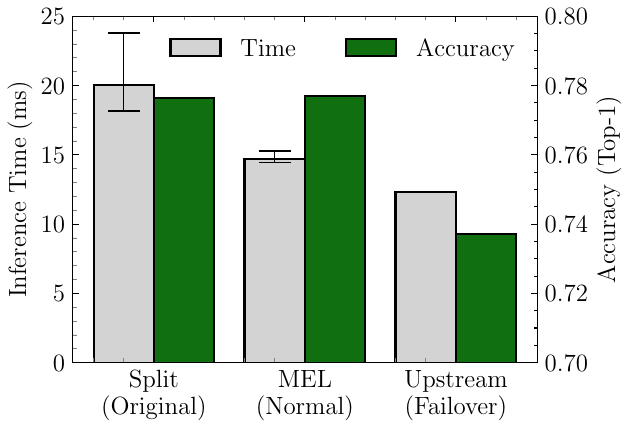}
  \caption{EfficientNet-B0 ($\Bc$5) response time on CIFAR-100.}
  \label{fig:respone_time}
    \end{minipage}
    \hfill\hfill
\end{figure}

DNNs are notorious for their accuracy-size trade-offs, where resource availability dictates the accuracy. 
\autoref{fig:size_effbet} show the effect of the ensemble size by comparing the number of parameters in  $h_{\{\mathcal{S}\}}$ to the parameters in the \emph{original} model on the accuracy of $h_{\{1\}}$, $h_{\{2\}}$, and $h_{\{1,2\}}$. The results highlight that as the size of the ensemble approaches 40\% of the original model, $\Bc5$ in this case, the performance of the \systemName ensemble becomes similar to that of the original model. Moreover, the results show that increasing the number of blocks raises the total size, even surpassing the original model, with diminishing returns, highlighting considerations for users when choosing their backup models.

\input{tables/hierarchical}

\subsubsection*{Hierarchical Training.} As discussed in ~\autoref{sec:system}, a key design decision for the upstream models is the ability to adjust the model complexity by changing the output granularity,  \emph{i.e.} utilizing the 20 superclasses coarse labels from CIFAR-100 to train upstream models $h_{\{1\}}$, $h_{\{2\}}$, while training downstream $h_{\{1,2\}}$ on the fine labels of the dataset. The effect of such a learning procedure is shown in ~\autoref{tab:hier}. Notably, the smaller backup models have 83.6\% accuracy, attributed to the fact that the subproblem is less complex, and the ensemble accuracy is still maintained on the fine-label classes.

\subsubsection*{Finetuning. } ~\autoref{tab:hier} depicts the effect of the additional fine-tuning on the downstream model while freezing the weights of the upstream models. The results show that the fine-tuning step is more beneficial in a hierarchical learning design, as in the case of EfficientNet-B0 ($\Bc$4), increasing the accuracy from 71.88\% to 72.15\%.

\subsubsection*{Downstream model architecture.}
\autoref{tab:heads} compares the accuracy of \systemName across different downstream model architectures. We focus on two architectures for the downstream models, fully connected networks (FC) and convolutional neural networks (CNNs), and test different sizes. 
The results show that \emph{using different architectures for the downstream models can optimize the ensemble accuracy with little to no effect on the upstream models.} The table also show that increasing the size of the FC networks by adding hidden layers typically does not increase the accuracy. For instance, going from a concatenation layer and a classification layer (\emph{i.e.,}FC (None)) to adding a hidden layer with 256 neurons did not enhance the accuracy. We note that this issue can be resolved by using higher weights for the upstream models (see \autoref{tab:heads}).
In contrast, increasing the size of CNNs (\emph{i.e,} by adding more filters) typically increases the accuracy. Similarly, we note that higher weights for the upstream models can further enhance the accuracy(see \autoref{tab:heads}).
\input{tables/downstream-architecture}

\subsubsection*{Effect of relative importance.}
\autoref{tab:arxiv_app_weights} presents the effect of varying the relative importance of upstream and downstream models during MEL training with EfficientNet-B0 ($\Bc5$) on CIFAR-100, across different downstream head architectures and configurations. The results indicate that \emph{increasing the upstream model's relative importance (i.e. more $\lambda_1$ and $\lambda_2$ as compared to $\lambda_{12}$) typically improves overall accuracy}. For example, assigning a 2:1 weight in favor of the upstream models with the CNN (320) head yields a 1.82$\times$ improvement in downstream accuracy, while also enhancing the performance of the individual upstream models.
\input{tables/app-weights}

\subsection{Comparing to Baselines}\label{sec:eval_baselines}

\input{tables/comparing-compressed}

In this section, we evaluate the performance of \systemName in comparison with the baselines as outlined in \autoref{sub:baselines}. In addition to the original and small failover replica configurations, we introduce the \emph{individual} and \emph{standalone} baselines to explore different effects in these other natural alternatives to \systemName, as prior work is limited in addressing this specific problem setting.

\subsubsection*{Single failover replicas.}
Reasonably, we expect our upstream models $h_{\{1\}}$ and $h_{\{2\}}$ trained within the ensemble strategy to be as competitive as traditional compressed models. \autoref{tab:arxiv_smaller_models} draws this comparison among architectures and datasets between upstream models and a \emph{small} model of the same size trained in isolation, to mimic typical compact backup replicas. The results show that these models trained within the ensemble have proximate behaviors to their independently trained counterparts, which demonstrates the effectiveness of our training methodology in not compromising the individual effectiveness of the upstream backup models.
\input{tables/same-strawman}
\subsubsection*{Training strategies.}
\autoref{tab:arxiv_same_strawman} shows the performance of $h_{\mathcal{S}}$ architecture trained using our proposed \systemName ensembling approach against previously discussed baselines --- \textbf{standalone} (trained as a whole for one downstream accuracy), and the \textbf{individually-trained} ensemble (i.e., upstream and downstream models are trained in two stages). In almost all cases, the results show that our ensembling approach yields the highest accuracy compared to the baselines, where the results show that our training strategy allows the models to achieve resiliency by operating as smaller models while maintaining the accuracy of a standalone model, trained on a single objective. More importantly, although the individually trained models also offer resilience, our \systemName ensembles consistently achieve higher performance, highlighting the benefits of \systemName that produces mutually refining upstream models.
Additional comprehensive evaluation will also be available in \autoref{app:baselines}.

\begin{figure}[t]
  \centering
  \hfill
  \begin{subfigure}[b]{0.32\textwidth}
    \includegraphics[width=\textwidth]{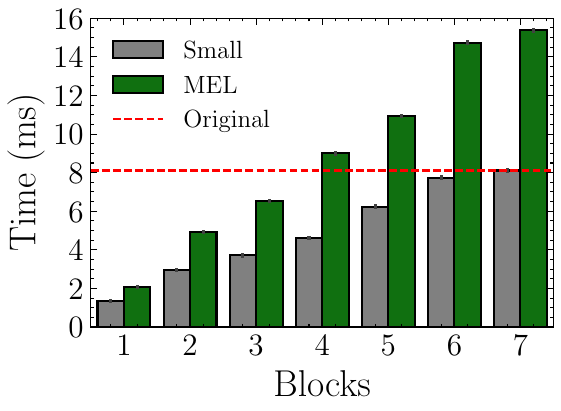}
    \caption{CPU}
    \label{fig:local_EENetB0_FC_CPU}
  \end{subfigure}
  \hfill
  \begin{subfigure}[b]{0.32\textwidth}
    \includegraphics[width=\textwidth]{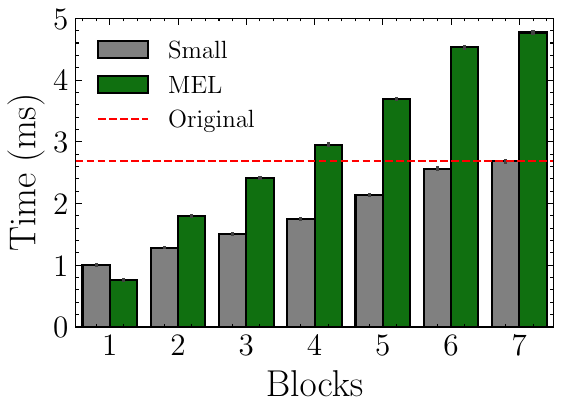}
    \caption{GPU}
    \label{fig:app_local_EENetB0_FC_GPU}
  \end{subfigure}
  \hfill
  \begin{subfigure}[b]{0.32\textwidth}
    \includegraphics[width=\textwidth]{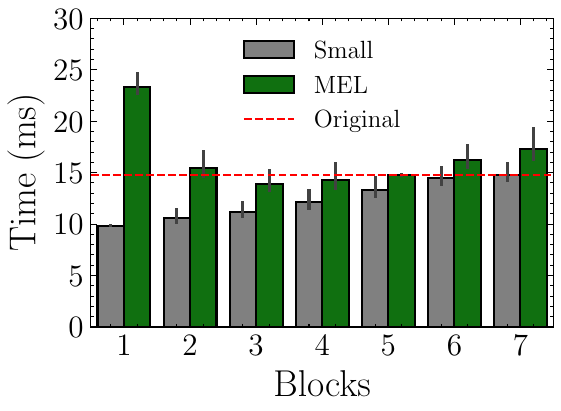}
    \caption{gRPC (GPU)}
    \label{fig:app_EENETB0_FC_rpc}
  \end{subfigure}
  \hfill
  \hfill
  \caption{Processing Latency of EfficientNet-B0 on CIFAR-100.}
  \label{fig:app_local_EENetB0_FC}
\end{figure}

\subsection{System Evaluation}
\autoref{fig:respone_time} shows the response time when deploying our architecture on three Nvidia A2 GPUs. 
We compare the accuracy and inference time of our \systemName based on EfficientNet-B0 ($\Bc$5) in the normal operation mode (\emph{i.e.,} all three servers are available), and a scenario where only an upstream model is available. For comparison, use a split inference approach~\cite{Matsubara2022:SplitSurvery} running the original models, where the model is split ($\Bc$1-5, $\Bc$6, and $\Bc$7/Classifier).
As shown, our approach demonstrates a lower inference time than a typical split model, as we are able to parallelize the execution of the two models by placing them on different servers, yielding a 25\% improvement in response time. 

We also evaluate the performance of the proposed ensemble models on both CPU and GPU, when deployed on single server for EfficientNet-B0, as in \autoref{fig:app_local_EENetB0_FC}. As one would expect, increasing the number of blocks increases latency — moving from 1 to 5 blocks increases latency by 4.8ms and 8.9ms for the standalone and \systemName models, respectively due to the fact that ensemble models are wider and require more computation. However, there is also more important aspect that when deployed across multiple servers (\emph{i.e.} in the intended setup), the results demonstrate that while using a smaller sub-blocks configuration may have lower parameter count, they might yield a higher response times -- consistent with prior split processing work \cite{Huang2020:CLIO}. Once past the initial layers, performance stabilizes, and \systemName introduces only minimal overhead. We expand our evaluation across scenarios in \autoref{app:systems}.

%% file: tables/datasets.tex
% \usepackage{multirow}
% \usepackage{graphicx}
\begin{table}[t]
\caption{Summary of Evaluated Datasets}
\centering
\label{tab:arxiv_datasets}
\resizebox{0.8\textwidth}{!}{
\begin{tabular}{cccc}
\toprule
Task & Dataset & Description & Coarse Labels \\
\toprule
\multirow{2}{*}{Image Classification} & CIFAR 100~\cite{krizhevsky2009:CIFAR} & 60K 32$\times$32 images & True (20 super classes) \\ 
& Tiered-ImageNet~\cite{Ren2018:Multi-Level} & 774K (subset of ImageNet-1k~\cite{Deng2009:imagenet}) & True (34 super classes) \\ 
 \midrule
Audio Classification, & \multirow{2}{*}{Speech Commands~\cite{speechcommandsv2}} & \multirow{2}{*}{65K (1-sec) recordings at 16kHz} & \multirow{2}{*}{False} \\      Speech recognition & & & \\ 
\midrule
LLM Pre-Training & BookCorpus~\cite{Zhu2015:Books} & 11K novels comprising of 1.2B tokens & False \\
\bottomrule
\end{tabular}
}
\vspace{-.3cm}
\end{table}

%% file: tables/results-intro.tex
\begin{table}[t]
\centering
\scriptsize
\caption{The ensemble model $h_{\{1,2\}}$ matches the original model's accuracy score on CIFAR-100, while preserving robustness and individual performance of $h_{\{1\}}$ and $h_{\{2\}}$.} %\todo{Clarify the parameters for h12 is inclusive of everything}
\label{tab:intro_results}
% \resizebox{\textwidth}{!}{%
\begin{tabular}{ccrcrcrcr}
\toprule
\multirow{3}{*}{DNN Model} & \multicolumn{6}{c}{\systemName} & \multicolumn{2}{c}{\multirow{2}{*}{Original}} \\
\cmidrule{2-7}
 & \multicolumn{2}{c}{$h_{\{1\}}$} & \multicolumn{2}{c}{$h_{\{2\}}$} & \multicolumn{2}{c}{$h_{\{1,2\}}$}\\ 
%\cmidrule{3-1}
& Acc. & Para. & Acc. & Para. & Acc. & Para. & Acc. & Para. \\ 
\midrule
% EfficientNet-B0 ($\Bc$3) & 0.5836 & 0.06M & 0.5902 & 0.06M &  \textbf{0.6261} & 0.2M & \multirow{2}{*}{0.7763} & \multirow{2}{*}{5.0M} \\ 
EfficientNet-B0 ($\Bc$5) & 0.7371 & 0.88M & 0.7428 & 0.88M &  {0.7769} & 1.79M & 0.7763 & 4.79M\\ 
Resnet50 ($\Bc$3) & 0.7324 & 8.76M & 0.7043 & 8.76M & {0.7552} & 17.50M & 0.7593 & 23.93M \\
\bottomrule
\end{tabular}
% }
\end{table}

%% file: tables/models-datasets.tex
\begin{table}[htbp]
\centering
\scriptsize
% \caption{Accuracy results across different data sets and architectures.}
\caption{\systemName performance is evaluated across datasets and DNN models. For each model, the task-relevant performance for individual models $h_{\{1\}}$ and $h_{\{2\}}$ and their combination $h_{\{1,2\}}$ is presented. {$^\dagger$Top-1 Acc. is reported for classification tasks (CIFAR-100, TieredImageNet, Speech Commands); Word Error Rate (WER) is reported for DeepSpeech2; and Perplexity is reported for BookCorpus.}} 
\label{tab:arxiv_generalization}
\begin{tabular}{crcrcrcrcr}
\toprule
\multirow{3}{*}{Dataset} & \multirow{3}{*}{DNN Model} &  \multicolumn{6}{c}{\systemName} & \multicolumn{2}{c}{\multirow{2}{*}{Original}} \\
\cmidrule{3-8}
& & \multicolumn{2}{c}{$h_{\{1\}}$} & \multicolumn{2}{c}{$h_{\{2\}}$}  & \multicolumn{2}{c}{$h_{\{1,2\}}$}  & &  \\ 
& & Perf.$^\dagger$ & Para. & Perf.$^\dagger$ & Para. & Perf.$^\dagger$ & Para. & Perf.$^\dagger$ & Para.\\
\midrule
\multirow{15}{*}{CIFAR-100} 
& EfficientNet-B0 ($\Bc$1) & 0.1883 & 4.08K & 0.2162 & 4.08K & 0.2365 & 11.45K & \multirow{7}{*}{0.7763} & \multirow{7}{*}{4.79M} \\
& EfficientNet-B0 ($\Bc$2) & 0.4070 & 21.59K & 0.4362 & 21.59K & 0.4634 & 48.08K \\
& EfficientNet-B0 ($\Bc$3) & 0.5824 & 0.07M & 0.5885 & 0.07M & 0.6261 & 0.15M \\
& EfficientNet-B0 ($\Bc$4) & 0.6785 & 0.32M & 0.6804 & 0.32M & 0.7169 & 0.65M \\
& EfficientNet-B0 ($\Bc$5) & 0.7371 & 0.88M & 0.7428 & 0.88M & \textbf{0.7769} & 1.79M \\
& EfficientNet-B0 ($\Bc$6) & 0.7529 & 2.90M & 0.7548 & 2.90M & \textbf{0.7888} & 5.83M \\
& EfficientNet-B0 ($\Bc$7) & 0.7308 & 3.60M & 0.7567 & 3.60M & \textbf{0.7838} & 7.3M\\
\cmidrule{2-10}
%\todo{Resnet 50 B1}
& Resnet50 ($\Bc$2)  & 0.6523 & 1.40M & 0.6488 & 1.40M & 0.6820 & 3.30M & \multirow{2}{*}{0.7593} & \multirow{2}{*}{23.93M} \\
& Resnet50 ($\Bc$3)  & 0.7324 & 8.76M & 0.7043 & 8.76M & 0.7552 & 17.50M & \\
\cmidrule{2-10}
& ViT-B/16 ($\Bc$1)  & 0.4644 & 7.80M & 0.4667 & 7.80M & 0.5066 & 15.90M & \multirow{6}{*}{0.6164} & \multirow{6}{*}{85.85M}\\
& ViT-B/16 ($\Bc$2)  & 0.5583 & 14.90M & 0.5652 & 14.90M & 0.5937 & 30.00M &\\
& ViT-B/16 ($\Bc$3)  & 0.5744 & 22.00M & 0.5860 & 22.00M & 0.6103 & 44.30M &\\
& ViT-B/16 ($\Bc$4)  & 0.5892 & 29.17M & 0.5908 & 29.17M & \textbf{0.6167} & 58.49M &  \\
& ViT-B/16 ($\Bc$5)  & 0.5909 & 36.00M & 0.5960 & 36.00M & \textbf{0.6250} & 72.00M & \\
& ViT-B/16 ($\Bc$6)  & 0.6030 & 43.00M & 0.6034 & 43.00M & \textbf{0.6295} & 86.00M &\\
                                 
\midrule
\multirow{3}{*}{TieredImageNet} 
& EfficientNet-B0 ($\Bc$4) & 0.6007 & 0.32M & 0.6048 & 0.32M & 0.6430 & 0.69M & \multirow{3}{*}{0.7352} & \multirow{3}{*}{4.79M} \\
& EfficientNet-B0 ($\Bc$5) & 0.6367 & 0.99M & 0.6359 & 0.99M & 0.6771 & 1.98M \\
& EfficientNet-B0 ($\Bc$6) & 0.6957 & 3.11M & 0.7096 & 3.11M & \textbf{0.7420} & 6.22M &\\
% &\todo{effnet b1-b5, b2-b5}
% &\todo{Resnet101-b3}
\midrule
\multirow{7}{*}{Speech Commands} 
& EfficientNet-B0 ($\Bc$2)   & 0.8042 & 0.02M & 0.8224 & 0.02M & 0.8413 & 0.05M & \multirow{5}{*}{0.9638} & \multirow{5}{*}{4.05M} \\
& EfficientNet-B0 ($\Bc$3)   & 0.9397 & 0.06M & 0.9379 & 0.06M & 0.9516 & 0.14M \\                        
& EfficientNet-B0 ($\Bc$4)   & 0.9587 & 0.31M & 0.9609 & 0.31M & \textbf{0.9643} & 0.64M & \\
& EfficientNet-B0 ($\Bc$5)   & 0.9632 & 0.86M & \textbf{0.9670} & 0.86M & \textbf{0.9690} & 1.74M &\\
& EfficientNet-B0 ($\Bc$6)   & 0.9596 & 2.90M & \textbf{0.9644} & 2.90M & \textbf{0.9686} & 5.81M &\\
\cmidrule{2-10}
& DeepSpeech2 ($\Bc$1) & 0.1590 & 7.27M & 0.1660 & 7.27M & 0.1412 & 14.58M & \multirow{2}{*}{0.1046}& \multirow{2}{*}{19.89M} \\
& DeepSpeech2 ($\Bc$2) & 0.1260 & 10.43M & 0.1270 & 10.43M & 0.1098 & 20.88M & \\
\midrule
\multirow{2}{*}{BookCorpus}     & GPT-mini ($\Bc$1) & 33.38 & 7.5M & 33.38 & 7.5M & 32.28 & 31.4M &  \multirow{2}{*}{23.50} & \multirow{2}{*}{33.6M} \\
                                & GPT-mini ($\Bc$2) & 28.16 & 10.5M & 28.02 & 10.5M & 27.38 & 37.7M & \\
                                 
\bottomrule
% \multicolumn{10}{l}{$^\dagger$Top-1 Acc. is reported for classification tasks (CIFAR-100, TieredImageNet, Speech Commands); Word Error Rate (WER) is reported}\\
% \multicolumn{10}{l}{for DeepSpeech2, and perplexity is reported for BookCorpus.}
\end{tabular}

% \todo{highlight accuracy metric.}

\end{table}

%% file: tables/hierarchical.tex
\begin{table}[t]
\scriptsize
\centering
\caption{Effect of hierarchical training and fine-tuning. Reducing task complexity improves the accuracy of \textit{small} models ($h_{\{1\}}, h_{\{2\}}$), while fine-tuning $h_{\{1,2\}}$ provides additional gains. Note that models trained on different granularities are incomparable, as they solve slightly different problems.}
\label{tab:hier}
\resizebox{\textwidth}{!}{%
\begin{tabular}{cccccccccc}
\toprule
\multirow{2}{*}{Dataset} & \multirow{2}{*}{DNN Model}   & \multicolumn{4}{c}{Fine-Grain Labels} & \multicolumn{4}{c}{Coarse-Grain Labels} \\
\cmidrule{3-10}
& & $h_{\{1\}}$ & $h_{\{2\}}$ & $h_{\{1,2\}}$ w/o FT & $h_{\{1,2\}}$ & $h_{\{1\}}$ & $h_{\{2\}}$ & $h_{\{1,2\}}$ w/o FT & $h_{\{1,2\}}$\\
\midrule
\multirow{3}{*}{CIFAR-100} 
%& EfficientNet-B0 ($\Bc$3) & n1 & n2 & n12 & n12 & 0.6966 & 0.7029 & 0.6015 & 0.6033  \\
    & EfficientNet-B0 ($\Bc$4) & 0.6785 & 0.6804 & 0.7147 & 0.7169 & 0.7890 & 0.8104 & 0.7188 & 0.7215 \\
    & EfficientNet-B0 ($\Bc$5) & 0.7371 & 0.7428 & 0.7765 & 0.7769 & 0.8197 & 0.8360 & 0.7610 & 0.7617  \\
%& Resnet50 ($\Bc$2) &n1 & n2 & n12 & n12 & 0.7013 & 0.7121 & 0.6485 & 0.6516\\
    & Resnet50 ($\Bc$3) & 0.7324 & 0.7043 & 0.7512 & 0.7552 & 0.7704 & 0.8005 & 0.7402 & 0.7423 \\
\bottomrule
\end{tabular}
}
\end{table}

%% file: tables/downstream-architecture.tex
\begin{table}[t]
\scriptsize
\centering
\caption{Comparison of different downstream architectures in MEL for combining upstream outputs. FC (None) uses only a classification layer; FC (256) adds a dense layer with 256 units. CNN (160) and CNN (320) include convolutional layers with the specified number of channels before classification.}
\label{tab:heads}
% \resizebox{\textwidth}{!}{%
\begin{tabular}{ccccccc}
\toprule
\multirow{2}{*}{Dataset} & \multirow{2}{*}{DNN Model}   & \multirow{2}{*}{Downstream Model (Size)} & \multicolumn{3}{c}{MEL} & \multirow{2}{*}{Total Size} \\
\cmidrule{4-6}
& & & $h_{\{1\}}$ & $h_{\{2\}}$ & $h_{\{1,2\}}$ &\\
\midrule
\multirow{8}{*}{CIFAR-100} 
& \multirow{4}{*}{EfficientNet-B0 ($\Bc$5)} & FC (None) & 0.7371 & 0.7428 & 0.7769 & 1.79M\\
& & FC (256) & 0.7224 & 0.7412 & 0.7610 & 1.81M \\
& & CNN (160) & 0.7156 & 0.7523 & 0.7743 & 1.79M \\
& & CNN (320) & 0.7451 & 0.7467 & \textbf{0.7810} & 1.83M \\
\cmidrule{2-7}
& \multirow{4}{*}{Resnet50 ($\Bc$3)} & FC (None) & 0.7324 & 0.70432 & 0.7552 & 17.50M\\
& & FC (256) & 0.7113 & 0.7252 & 0.6931 & 17.84M \\
& & CNN (160) & 0.7060 & 0.7223 & 0.7483 & 17.64M \\
& & CNN (320) & 0.7058 & 0.7161 & \textbf{0.7557} & 17.98M \\
% &\todo{effnet b1-b5, b2-b5} \\
% &\todo{Resnet101-b3} \\
\midrule
\multirow{4}{*}{TieredImageNet}
& \multirow{4}{*}{EfficientNet-B0 ($\Bc$5)} & FC (None) & 0.6367 & 0.6359 & 0.6771 & 1.98M\\
& & FC (256) & 0.6492 & 0.6560 & 0.6713 & 2.05M \\
& & CNN (160) & 0.6574 & 0.6595 & 0.6873 & 1.98M \\
& & CNN (320) & 0.6618 & 0.6665 & \textbf{0.7022} & 2.11M \\
\bottomrule
\end{tabular}
% }
\end{table}

%% file: tables/app-weights.tex
\begin{table}[t]
\scriptsize
\centering
\caption{Impact of varying weights on the upstream vs downstream models during MEL training with EfficientNet-B0 ($\Bc$5) on CIFAR-100. $\lambda_1$ corresponds to the weight of the upstream models (\emph{i.e.} $h_{\{1\}}$, $h_{\{2\}}$), while $\lambda_{12}$ controls the weight of the downstream model $h_{\{1,2\}}$ in the loss formulation. Top-1 accuracy is reported across the different downstream head architectures. We omit the fine-tuning step to show the effect of weights on the accuracy of all models.}
\label{tab:arxiv_app_weights}
% \resizebox{\textwidth}{!}{%
\begin{tabular}{ccccc}
\toprule
\multirow{2}{*}{Downstream Model (Size)} & \multirow{2}{*}{$\lambda_{1} : \lambda_{12}$} & \multicolumn{3}{c}{MEL} \\
\cmidrule{3-5}
& & $h_{\{1\}}$ & $h_{\{2\}}$ & $h_{\{1,2\}}$\\
\midrule
\multirow{4}{*}{FC (None)}
& 1\, : \,5 & 0.6317 & 0.7325 & 0.7631\\
& 1\, : \,2 & 0.7064 & 0.7484 & 0.7735\\
& 1\, : \,1 & 0.7371 & 0.7428 & 0.7769\\
& 2\, : \,1 & 0.7267 & 0.7522 & 0.7754\\
& 5\, : \,1 & 0.7515 & 0.7528 & \textbf{0.7787}\\
\midrule
\multirow{4}{*}{FC (256)}
& 1\, : \,5 & 0.5687 & 0.7238 & 0.7479\\
& 1\, : \,2 & 0.7164 & 0.7458 & 0.7665\\
& 1\, : \,1 & 0.7224 & 0.7412 & 0.7610\\
& 2\, : \,1 & 0.7464 & 0.7631 & 0.7840\\
& 5\, : \,1 & 0.7536 & 0.7654 & \textbf{0.7852}\\
\midrule
\multirow{4}{*}{CNN (160)}
& 1\, : \,5 & 0.6629 & 0.7391 & 0.7764\\
& 1\, : \,2 & 0.6986 & 0.7617 & 0.7825\\
& 1\, : \,1 & 0.7156 & 0.7573 & 0.7743\\
& 2\, : \,1 & 0.7571 & 0.7661 & \textbf{0.7882}\\
& 5\, : \,1 & 0.7611 & 0.7625 & 0.7832\\
\midrule
\multirow{4}{*}{CNN (320)}
& 1\, : \,5 & 0.6966 & 0.7032 & 0.7735\\
& 1\, : \,2 & 0.7325 & 0.7493 & 0.7814\\
& 1\, : \,1 & 0.7451 & 0.7467 & 0.7810\\
& 2\, : \,1 & 0.7602 & 0.7609 & \textbf{0.7917}\\
& 5\, : \,1 & 0.7600 & 0.7625 & 0.7883\\
\bottomrule
\end{tabular}
% }
\end{table}

%% file: tables/comparing-compressed.tex
\begin{table}[t]
\centering
\scriptsize
\caption{Performance of individual backup models $h_{\{1\}}$, $h_{\{2\}}$ under joint optimization within the MEL framework compared to individual training.}
\label{tab:arxiv_smaller_models}
\begin{tabular}{ccccc}
\toprule
\multirow{2}{*}{Dataset} & \multirow{2}{*}{DNN Model} & \multicolumn{2}{c}{\systemName Perf.} & \multirow{2}{*}{Standalone} \\
\cmidrule{3-4}
 &  & $h_{\{1\}}$ & $h_{\{2\}}$ & \\ 
\midrule
\multirow{3}{*}{CIFAR-100}  
& EfficientNet-B0 ($\Bc$1) & 0.1883 & \textbf{0.2162} & 0.2060 \\
& EfficientNet-B0 ($\Bc$2) & 0.4070 & \textbf{0.4362} & 0.4268 \\
& EfficientNet-B0 ($\Bc$3) & 0.5824 & \textbf{0.5885} & 0.5809 \\
& EfficientNet-B0 ($\Bc$4) & 0.6785 & 0.6804 & 0.7105 \\
& EfficientNet-B0 ($\Bc$5) & 0.7428 & 0.7371 & 0.7471 \\
& EfficientNet-B0 ($\Bc$6) & 0.7529 & 0.7548 & 0.7686 \\
\cmidrule{2-5}
& Resnet50 ($\Bc$2) & 0.6523 & 0.6488 & 0.6801  \\
& Resnet50 ($\Bc$3) & \textbf{0.7324} & 0.7043 & 0.7263  \\
\cmidrule{2-5}
& ViT-B/16 ($\Bc$1) & 0.4644 & 0.4667 & 0.4740\\
& ViT-B/16 ($\Bc$2) & 0.5583 & 0.5652 & 0.5685\\
& ViT-B/16 ($\Bc$3) & 0.5744 & 0.5860 & 0.5916\\
& ViT-B/16 ($\Bc$4) & 0.5892 & 0.5908 & 0.6013\\
& ViT-B/16 ($\Bc$5) & 0.5909 & 0.5960 & 0.6044\\
& ViT-B/16 ($\Bc$6) & 0.6030 & 0.6034 & 0.6100\\
\midrule
\multirow{1}{*}{Tiered-Imagenet} 
& EfficientNet-B0 ($\Bc$4) &  0.6007 & \textbf{0.6048} & 0.5847 \\
& EfficientNet-B0 ($\Bc$5) &  0.6367 & 0.6359 & 0.6555 \\
& EfficientNet-B0 ($\Bc$6) &  0.6957 & \textbf{0.7096} & 0.6905 \\
\midrule
\multirow{2}{*}{Speech Commands} 
& EfficientNet-B0 ($\Bc$2) & 0.8042 & \textbf{0.8224} & 0.8134 \\
& EfficientNet-B0 ($\Bc$3) & 0.9397 & 0.9379 & 0.9421 \\
& EfficientNet-B0 ($\Bc$4) & 0.9587 & 0.9609  & 0.9625 \\
\cmidrule{2-5}
& DeepSpeech2 ($\Bc$1)  & 0.1590 & 0.1660  & 0.1520 \\
& DeepSpeech2 ($\Bc$2)  & 0.1260 & 0.1270  & 0.1185 \\
\midrule
\multirow{2}{*}{BookCorpus}
& GPT-mini ($\Bc$1) & \textbf{33.38} & 33.38 & 33.41 \\
& GPT-mini ($\Bc$2) & 28.16 & \textbf{28.02} & 28.67 \\

% \midrule
% \multirow{1}{*}{Bookcorpus}         
%     & GPT-mini ($\Bc$1) & 3.5080 & 3.508 & 3.497 \\
                                 
\bottomrule
\end{tabular}
\end{table}

%% file: tables/same-strawman.tex
\begin{table}[t]
\centering
\scriptsize
\caption{Performance of $h_{\{1,2\}}$ under different training strategies --{Standalone} trains only $h_{\{1,2\}}$, while Individually trained trains $h_{\{1\}}, h_{\{2\}}$ independently and then optimizes $h_{\{1,2\}}$ with frozen upstream models.}

\label{tab:arxiv_same_strawman}
\begin{tabular}{ccccc}
\toprule
Dataset & DNN Model & \systemName Perf. & Standalone & Ind. Trained \\
\midrule
\multirow{3}{*}{CIFAR-100}  
    & EfficientNet-B0 ($\Bc$1) & 0.2365 & 0.2523 & 0.2086 \\
    & EfficientNet-B0 ($\Bc$2) & 0.4634 & 0.4662 & 0.4155 \\
    & EfficientNet-B0 ($\Bc$3) & \textbf{0.6261} & 0.6181 & 0.6072 \\
    & EfficientNet-B0 ($\Bc$4) & \textbf{0.7169} & 0.7064 & 0.7038 \\
    & EfficientNet-B0 ($\Bc$5) & \textbf{0.7769} & 0.7524 & 0.7462 \\
    & EfficientNet-B0 ($\Bc$6) & \textbf{0.7888} & 0.7662 & 0.7615 \\
\cmidrule{2-5}
    & Resnet50 ($\Bc$2)& \textbf{0.6801} & 0.6338 & 0.6593\\
    & Resnet50 ($\Bc$3)& 0.7552 & 0.7579 & 0.7235 \\
\cmidrule{2-5}
    & ViT-B/16 ($\Bc$1)& \textbf{0.5066} & 0.5008 & 0.4829 \\
    & ViT-B/16 ($\Bc$2)& \textbf{0.5937} & 0.5874 & 0.5720 \\
    & ViT-B/16 ($\Bc$3)& 0.6103 & 0.6160 & 0.5947 \\
    & ViT-B/16 ($\Bc$4)& \textbf{0.6167} & 0.6111 & 0.6001 \\
    & ViT-B/16 ($\Bc$5)& 0.6250 & 0.6279 & 0.6071 \\
    & ViT-B/16 ($\Bc$6)& \textbf{0.6295} & 0.6149 & 0.6126 \\
\midrule
\multirow{1}{*}{Tiered-Imagenet} 
    & EfficientNet-B0 ($\Bc$5) & \textbf{0.6430} & 0.6365 & 0.5905 \\
    & EfficientNet-B0 ($\Bc$5) & 0.6771 & 0.6955 & 0.6589 \\
    & EfficientNet-B0 ($\Bc$6) & \textbf{0.7420} & 0.7207 & 0.6925 \\
\midrule
\multirow{2}{*}{Speech Commands} 
    & EfficientNet-B0 ($\Bc$2) & \textbf{0.8413} & 0.8495 & 0.8206 \\
    & EfficientNet-B0 ($\Bc$3) & \textbf{0.9516} & 0.9504 & 0.9457 \\
    & EfficientNet-B0 ($\Bc$4) & \textbf{0.9643} & 0.9623 & 0.9634 \\
    \cmidrule{2-5}
    & DeepSpeech2 ($\Bc$1)  & \textbf{0.1412} & 0.1546 & 0.1499  \\
    & DeepSpeech2 ($\Bc$2)  & \textbf{0.1098} & 0.1217 & 0.1181  \\
\midrule
\multirow{1}{*}{Bookcorpus}
    & GPT-mini ($\Bc$1) & 32.28 & 28.99 & 33.04 \\
    & GPT-mini ($\Bc$2) & 27.38 & 25.21 & 28.15 \\
                                 
\bottomrule
\end{tabular}
\end{table}

%% file: sections/5-conclusion.tex
In this paper, we presented our Multi-level Ensemble Learning (\systemName) approach for resilient edge inference and experimentally demonstrated its efficacy through extensive evaluation. By training ensemble backup models under joint optimization, it is able to take advantage of the fact that when multiple models succeed, the performance is mutually refined. Our work has a few limitations. First, in scenarios where edge resources are heterogeneous or utilized unevenly, the upstream models may need to be heterogeneous themselves (in terms of size), an aspect which we did not study. Secondly, the question of choosing the optimal configuration from a family of \systemName models that we trained is not explored in this work. Both limitations are also opportunities for future work. In terms of broader impact, our work promises a foundation for a robust edge AI to become more prevalent in the future and enable a new generation of resilient, latency-sensitive inference applications such as AR, personal robotics, and more.
%\lipsum[1]

%% file: sections/appendix.tex
The appendix of this paper is structured as follows.
\begin{itemize}
    \item \autoref{app:proof} presents the proof of Proposition~\ref{thm}.
    \item \autoref{app:more_system_details} presents the system design and implementation. 
    \item \autoref{app:exp_details} lists the details of the utilized datasets, model architecture, and computing resources.
    \item \autoref{app:more_models} expands the results of \systemName across models and architectures.
    \item \autoref{app:ablation} expands the ablation studies of \systemName across more upstream tasks, downstream architectures, and weights.
    \item \autoref{app:baselines} expands the baseline comparisons for \systemName.
    \item \autoref{app:systems} expands the performance evaluation comparisons for \systemName.
    \item  \autoref{app:limitations} presents the limitations of \systemName.
\end{itemize}

\begin{figure}[t!]
    \centering
    \includegraphics[width=\linewidth]{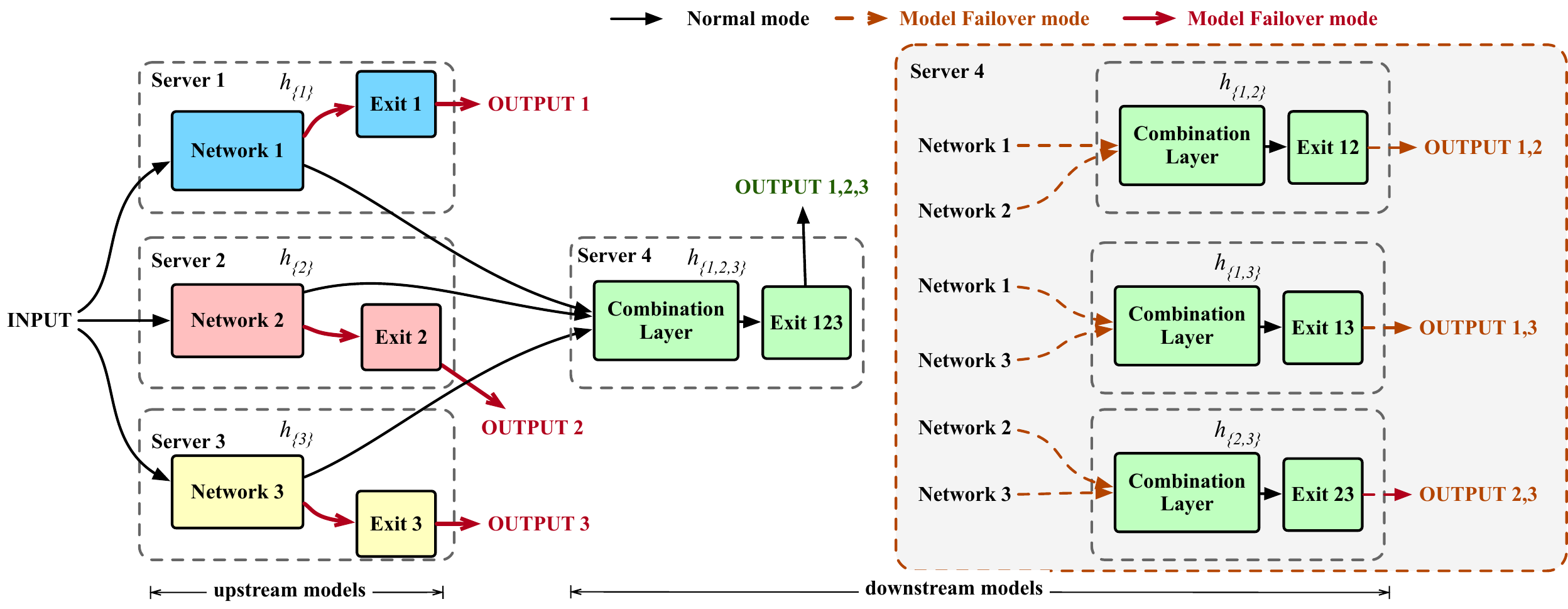}
    \caption{An example \emph{multi-level ensemble} architecture featuring three upstream models. The figure highlights that three upstream models require four downstream models: A downstream model for the normal model, where all three models are available, and three models for different failure scenarios. Lastly, the figure highlights that models can also have their own exits in case only one server is available.}
    \label{fig:ensemble_3_models}
\end{figure}

\section{Proof of Proposition~\ref{thm}}\label{app:proof}
\input{sections/app-B-proof}

\section{System Design and Implementation}\label{app:more_system_details}
\input{sections/app-C-system-design}

\section{Experimental Setup}\label{app:exp_details}
\input{sections/app-D-exp-details}

\section{Additional Evaluation Results}\label{app:more_models}
\input{sections/app-E-results}

\section{Additional Ablation Studies}\label{app:ablation}
\input{sections/app-F-ablation}

\section{More Baselines}\label{app:baselines}
\input{sections/app-G-baselines}

\section{More systems evaluation}\label{app:systems}
\input{sections/app-H-systems}

\section{Limitations}\label{app:limitations}
The main limitation of our work is that the number of models for the downstream tasks grows as $O(2^{\texttt{upstream models}})$ of the number of upstream models. 
For example, the ensemble with three upstream models (see \autoref{fig:ensemble_3_models}) requires 4 number of downstream models to handle different scenarios. One approach could be to train models using just one (or a limited number of) downstream models capable of executing without requiring all the representations from the upstream models. This could be achieved using a masking technique (\emph{e.g.,} adding zeros for failed models) while training, however, such an approach requires a different training methodology and is left for future work. 

%% file: sections/app-B-proof.tex
\begin{proof}
 We consider overall generalization error $\text{gen}^2_{overall} = \mathbb{E}_{\mathcal{D},h_{\{1\}},h_{\{2\}},h_{\{1,2\}}}[L(h_{\{1,2\}},h_{\{1\}},h_{\{2\}})]=\mathbb{E}[\frac{1-p}{1+p}(L(h_{\{1,2\}})-\hat L (h_{\{1,2\}}))^2 + \frac{p}{1+p}(L(h_{\{1\}})-\hat L(h_{\{1\}}))^2+\frac{p}{1+p}(L(h_{\{2\}})-\hat L (h_{\{2\}})^2]$, where $0 \leq p\leq 1$ acts as a failover probability, i.e. when $p=1$ it is guaranteed to use a lightweight model, and $p=0$ it is guaranteed to use the ensemble model. Note that the overall generalization error (sum of squares) yields,
\begin{align*}
    \text{gen}^2_{overall} \leq \frac{1-p}{1+p}\frac{2\sigma^2}{n}I(\mathcal{D};h_{\{1,2\}})+\frac{p}{1+p}\frac{2\sigma^2}{n}(I(\mathcal{D};h_{\{1\}})+I(\mathcal{D},h_{\{2\}}))
\end{align*}
Using the above Markov chain and data processing inequality, we can show,
\begin{align}
    I(\mathcal{D},h_{\{1,2\}})\leq I(\mathcal{D};h_{\{1\}},h_{\{2\}}) = I(D;h_{\{1\}})+I(D;h_{\{2\}})-I(h_{\{1\}},h_{\{2\}}),
\end{align}
where we used the following identities given the conditional independence,
\begin{align*}
    I(D;h_{\{1\}},h_{\{2\}})&= H(h_{\{1\}},h_{\{2\}})-H(h_{\{1\}},h_{\{2\}}|D)\\
    &= [H(h_{\{1\}})+H(h_{\{2\}})-I(h_{\{1\}};h_{\{1,2\}})]-[H(h_{\{1\}}|D)]+H(h_{\{2\}}|D)\\
    &= [H(h_{\{1\}})-H(h_{\{1\}}|D)]+[H(h_{\{2\}})-H(h_{\{2\}}|D)]-I(h_{\{1\}};h_{\{2\}})\\
    &= I(D;h_{\{1\}})+I(D;h_{\{2\}})-I(h_{\{1\}};h_{\{2\}}).
\end{align*}
As a result, we have,
\begin{align*}
    \text{gen}^2_{overall} \leq \frac{1}{1+p}\frac{2\sigma^2}{n}((I(D;h_{\{1\}})+I(D;h_{\{2\}}))-(1-p)(I(h_{\{1\}};h_{\{2\}})))
\end{align*}
\end{proof}

%% file: sections/app-C-system-design.tex
\begin{figure}[t]
    \centering
    \includegraphics[width=0.6\linewidth]{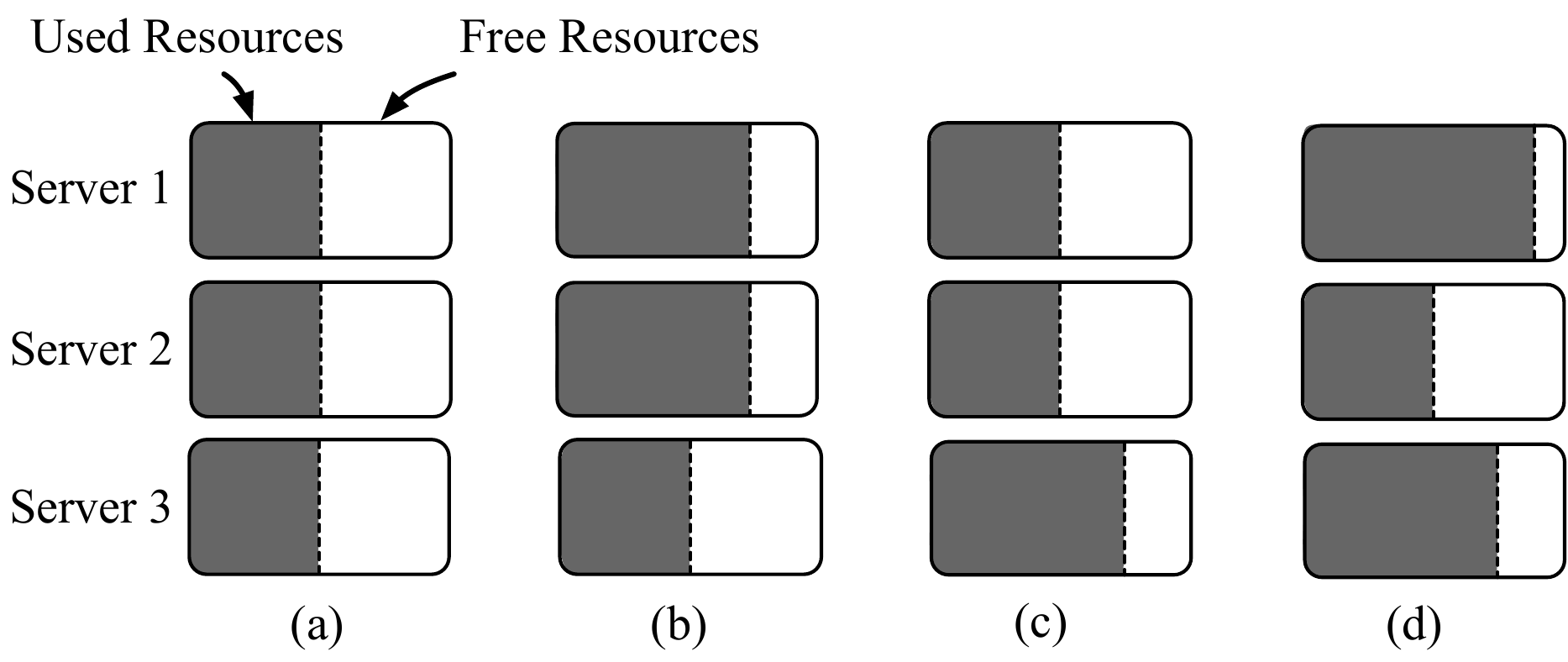}
    \caption{Resource fragmentation scenarios, where servers have equal available resources (a), and scenarios where one of the servers has more capacity than the others (b), and a scenario where one of the servers is more limited than the others (c), and (d) a scenario where all servers have different amount of resources.}
    \label{fig:app_freg}
\end{figure}

Although model selection has been extensively studied~\cite{Ahmad2024:Loki, Ahmad2024:Proteus, Hanafy2021:DNNSelection, wu2025:FailLite, Wan2020:ALERT}, none of the previous approaches have considered selection for model ensembles. To instantiate an ensemble model, as stated in ~\autoref{sec:system}, we must consider the available resources, which may be fragmented as we show in \autoref{fig:app_freg}.  

To select a model, we must consider the interplay between the design decisions and runtime considerations. Our design decisions include:
1) Size and number of upstream models, 2) Size and architecture of the upstream model, 3) Relative importance of the upstream and downstream models, and 4) Output granularity. Although our ensemble architectures work on the block level, where upstream models are a prefix of the models, which highly discretize the problem, the problem is still not trivial. Picking the right size upstream model involves navigating a trade-off between model size, processing latency, and accuracy. For instance, as shown in \autoref{fig:size_effbet} and \autoref{tab:arxiv_generalization}, using the first six blocks of the EfficientNetB0 barely adds any accuracy despite having 3.47$\times$ more parameters. However, picking small models may actually lead to worse latency, as shown in \autoref{fig:app_EENETB0_FC_rpc}, as earlier models have larger intermediate representations, an issue highlighted in earlier research~\cite {Huang2020:CLIO}. 
Additionally, deciding the size and relative importance of the downstream models is also nontrivial, as they are highly dependent on the available resources and failure type. For instance, as we show in \autoref{tab:heads}, picking a larger downstream model may not always increase the accuracy. Moreover, emphasizing training the accuracy of the downstream model by giving it higher relative importance may not yield higher accuracy, as we highlight in \autoref{tab:arxiv_app_weights}.

\subsection{Ensemble Model Selection}

Due to the complexity of ``right-sizing'' the model ensemble at training time, as the available resources and failure are hard to know beforehand. Our approach, similar to previous work~\cite{wu2025:FailLite, Ahmad2024:Proteus}, trains a family of multiple model ensembles and selects the appropriate model at runtime. 
\autoref{alg:algorithm_family} lists an exemplar heuristic that creates a family of trained ensembles. The algorithm takes the original architecture, possible architectures for the upstream model, and a resource budget. The resource budget can be a memory size or a latency SLO that the model must respect. To select the model family, we first iterate over the original model blocks and possible downstream architectures. We assume that the downstream architectures come in different sizes (e.g., number of filters in a CNN or number of hidden layers in a fully connected model). Next, we check if the selected number of blocks and the downstream architecture respect the resource budget. Lastly, the user can train these models and apply different granularities for models that have low accuracy. 
At runtime, systems can use a placement algorithm (e.g., a worst-fit heuristic). However, evaluating the performance of such runtime selection and placement approaches is part of our future work.

\RestyleAlgo{ruled}
\begin{algorithm}[t]
    \LinesNumbered
    \SetAlgoNoEnd
    \footnotesize
    \caption{\texttt{Ensemble\_Family()}}
    \label{alg:algorithm_family}
    \KwIn{Original model architecture $O$, Upstream models architectures $A$, Resource budget $\Rc$.}
    \KwOut{Ensemble Family $E$.} 
    %\tcp{Compute Available Capacity.}
    \textbf{Initialization:} $E \gets [ \ ]$\;
    \For(\tcp*[f]{Iterate over Blocks.}){$\Bc \in O.\texttt{Blocks}$}{
    \For(\tcp*[f]{Iterate over upstream architectures.}){$\Ac \in A$}{
        \If(\tcp*[f]{Ensembles must be within the resource budget.}){\texttt{feasible}($\Bc, \Ac, \Rc$)}{
        $E.\texttt{append}(\Bc, \Ac)$\;
        }
        }
    }
    \Return $E$
\end{algorithm}

\subsection{System Implementation} 
In this section, we detail the implementation of our failure-resilient model serving system. Although our multi-level model ensembles can be deployed on any AI inference systems, our current implementation is based on ONNX Runtime environment v1.21. After training the models using Pytorch v2.5.1, we export the sub-models into ONNX format and run them using the ONNX runtime. We export sub-models individually to different files and allow the users to send inference requests and for sub-models to exchange intermediate representation via a gRPC interface. The ONNX runtime allows for multiple execution providers. Our evaluation focuses on \texttt{CPUExecutionProvider} that allows running the models on CPUs and \texttt{ CUDAExecutionProvider} that allows the models to run on GPUs using CUDA and cuDNN implementations. Our model inference implementation, which we plan to open-source, is $\sim$1k SLOC. Note that our current evaluation is limited to inference latency across different architectures and ensemble sizes, evaluating failure resiliency metrics, such as mean time to recovery (MTTR), and strategies such as warm and cold backups are part of our future work.

%% file: sections/app-D-exp-details.tex
In this section, we detail our datasets, architectures, training configuration, and compute resources for training and inference.

\subsection{Datasets}
Our training methodology is a general training methodology that applies to any data. In addition to the datasets listed in ~\autoref{tab:arxiv_datasets}, we add the BookCorpus dataset \cite{Zhu2015:Books}.
Here we detail the four publicly available datasets we used.

\begin{itemize}
    \item \textbf{CIFAR-100~\cite{krizhevsky2009:CIFAR}:} The CIFAR-100 dataset contains 60k 32×32 images and 
     has 100 classes, grouped into 20 superclasses, \emph{e.g.,} the \texttt{fruit and vegetables} class has \texttt{apples, mushrooms, oranges, pears, sweet peppers}. We use this dataset for an image classification task, and report performance in terms of Top-1 accuracy. 
    \item \textbf{Tiered-ImageNet~\cite{Ren2018:Multi-Level}:} The Tiered-ImageNet dataset is a subset of ImageNet with 774k samples representing 608 classes, grouped into 34 higher-level nodes in the ImageNet human-curated hierarchy. However, we re-partitioned the training, validation, and test data to ensure that the model was exposed to all label classes, which was not a requirement in the original dataset construction. We use this dataset for an image classification task, and report performance in terms of Top-1 accuracy.
    \item \textbf{Speech Commands~\cite{speechcommandsv2}:} The Speech Commands dataset contains 65k (1-sec) recordings at 16kHz across 35 classes (\emph{i.e., words}). The dataset is used for classification and Speech Recognition, where we report Top-1 accuracy and Word Error Rate (WER), respectively.
    \item \textbf{BookCorpus~\cite{Zhu2015:Books}:} Lastly, the BookCorpus dataset is composed of 11k novels comprising 1.2B tokens (using a vocab size of 8000), and we use this dataset for LLM Pre-Training, where we report the model perplexity.   
\end{itemize}

\input{tables-app/app-architectures}
\subsection{Architectures} 
As highlighted in \autoref{sec:eval_setup}, we use several architectures to validate our training approach. \autoref{tab:app_arcs} lists the models and the architecture used for our evaluation. The table also highlights the number of blocks in the original model. Note that although there is no technical limit on the number of blocks, we limit the blocks based on the description of the original paper.
\subsection{Training Configurations}
In this section, we detail the training configurations, summarized in \autoref{tab:app_training_configs}.
For each dataset, the same configuration is shared between model sizes and configurations. 
As highlighted earlier, our ensemble models are trained in two steps. In the first step, we train the entire ensemble using a weighted loss function as explained in \autoref{sec:problem}. Then we also fine-tune the downstream model while freezing upstream models, which further improves the results as we highlight later in \autoref{tab:app_hier}.
In all the training and fine-tuning steps, we use the AdamW\cite{loshchilov2019:adamW} optimizer with the cosine schedules\cite{loshchilov2017:sgdr}.
\input{tables-app/app-training-configs}

\subsection{Training environment}
We train models in a local SLURM cluster with many GPU types.The detailed cluster configuration is (\emph{**url hidden for anonymity**})
Most of the training uses a single GPU except for classification on TieredImageNet and language modeling on BookCorpus, which is done using 2 and 4 GPU nodes respectively. The training is implemented using PyTorch v2.5.1 and is based on the TIMM\cite{rw2019timm} and other open-source implementations, which includes various architectures like ResNet, EfficientNet, and Vision Transformers. \systemName models of EfficientNet and Resnet takes 5-10 hours to train on a single GPU on CIFAR-100 depending on the size (i.e. sub-blocks in the model). Training Vision Transformer and training on the TieredImageNet takes much longer around 15-24 hours on multiple GPUs. We plan to open-source our training scripts.

\subsection{AI inference environment}
Our AI inference environment is based on a local cluster of four machines with three Nvidia A2 GPUs running CUDA v12.4 and cuDNN v9.8.0. The machines are Dell PowerEdge servers with  Intel(R) Xeon(R) CPU E5-2660 v3 @ 2.60GHz processor and connected via a 10 Gbps network. We evaluate the performance of our architecture by evaluating the local performance on a GPU and a CPU, as well as evaluating the model serving system by deploying the models across servers.

%% file: tables-app/app-architectures.tex
\begin{table}[t]
    \centering
        \caption{Models architectures and number of blocks in the original model.}
    \label{tab:app_arcs}
    \begin{tabular}{ccc}
    \toprule
      Model & Architecture & Blocks \\
    \toprule
       EfficientNet-B0 & CNN & 7\\
       % EfficientNet-B1 & CNN & 7\\
       % EfficientNetB2 & CNN & 7\\ \midrule
         ResNet50 & CNN & 4\\ 
         ViT-B/16 & Transformer & 12 \\ 
         DeepSpeech2 & GRU & 6\\
         GPT-mini & Transformer & 8\\
         \bottomrule
    \end{tabular}

\end{table}

%% file: tables-app/app-training-configs.tex
\begin{table}[t]
    \centering
    \small
    \caption{Training Configurations. BS is the batch size. LR is the learning rate reporting as starting learning rate with 10 warmup epochs and a cosine rate decay schedule. FT is a fine-tuning step for fine-tuning the downstream model. $^*$For BookCorpus, batch size is reported in tokens per update, and training/fine-tuning steps reflect total token count.}
    \label{tab:app_training_configs}
    \resizebox{\textwidth}{!}{%
    \begin{tabular}{clcccc}
    \toprule
        Dataset & Architectures & BS &Learning Rate & Training Epochs & FT Epochs \\\toprule
        \multirow{3}{*}{CIFAR-100} 
        &  EfficientNet-B0 (\emph{Small})  & 64 & 0.005 & 100 & - \\
        &  EfficientNet-B0 ($\Bc$1 - $\Bc$7)  & 64 & 0.005 & 150-200 & 50 \\
        &  EfficientNet-B0 (Three-upstream)  & 64 & 0.005 & 200 & 50 \\
        % &  EfficientNetB0 - Original & & & & \\\midrule
        &  Resnet50 (\emph{Small}) & 64 & 0.005 & 100 & - \\ 
        &  Resnet50 ($\Bc$1 - $\Bc$3) & 64 & 0.005 & 150-200 & 50 \\ 
        &  Resnet50 (Three-upstream) & 64 & 0.005 & 200 & 50 \\ 
        & ViT-B/16 (\emph{Small}) & 64 & 0.0001 & 150 & - \\
        & ViT-B/16 ($\Bc$1 - $\Bc$6) & 64 & 0.0001 & 150-250 & 50 \\
        & ViT-B/16 (Three upstream) & 64 & 0.0001 & 250 & 50 \\
        \midrule
        TieredImageNet
        & EfficientNet-B0 & 128 & 0.005 & 150-200 & 50\\
        % & EfficientNet-B1 & 128 & 0.005 & 150 & 50\\
        \midrule
        \multirow{2}{*}{Speech-commands}
        & EfficientNet-B0 & 64 & 0.001 & 100 & 30\\
        & DeepSpeech2 & 64 & 0.001 & 100 & 30\\
        \midrule
        BookCorpus & GPT-mini & 65000$^*$ & 0.0006 & 13.1 billion$^*$ & 1.6 billion$^*$ \\
        \bottomrule
    \end{tabular}
    }
    % \todo{fill and add details you see fit.}
\end{table}

%% file: sections/app-E-results.tex
\subsection{Hierarchical training across sizes}
\input{tables-app/app-hierarchical}

\autoref{tab:app_hier} expands our results in \autoref{tab:hier}. The table illustrates the effect of using coarse-grain labels as a design decision while training upstream models in the \systemName framework. As shown, \emph{reducing the task complexity helps create better performance across the upstream models, without compromising much performance on the downstream task, when the models are sufficiently big}. In some cases, hierarchical training even leads to better overall performance. For example, EfficientNet-B0 ($\Bc$5) on TieredImageNet yields a 1.4\% improvement in accuracy on the fine-labels while achieving 82.2\% performance on the upstream models. Similar results are also observed across multiple configurations of Vision Transformer models on CIFAR-100.

%% file: tables-app/app-hierarchical.tex
\begin{table}[t]
\scriptsize
\centering
\caption{Effect of hierarchical training. Reducing task complexity improves the accuracy of \textit{small} models ($h_{\{1\}}, h_{\{2\}}$). However, note that models trained on different granularities are not directly comparable, as they solve slightly different problems.}
\label{tab:app_hier}
% \resizebox{\textwidth}{!}{%
\begin{tabular}{cccccccc}
\toprule
\multirow{2}{*}{Dataset} & \multirow{2}{*}{DNN Model}   & \multicolumn{3}{c}{Fine-Grain Labels} & \multicolumn{3}{c}{Coarse-Grain Labels} \\
\cmidrule{3-8}
& & $h_{\{1\}}$ & $h_{\{2\}}$ & $h_{\{1,2\}}$ & $h_{\{1\}}$ & $h_{\{2\}}$ & $h_{\{1,2\}}$\\
\midrule
\multirow{3}{*}{CIFAR-100} 
& EfficientNet-B0 ($\Bc$3) & 0.5824 & 0.5885 & 0.7169 & 0.6966 & 0.7029 & 0.6033\\
\cmidrule{2-8}
& Resnet50 ($\Bc$2) &0.6523 & 0.6488 & 0.6820 & 0.7013 & 0.7121 & 0.6516\\
\cmidrule{2-8}
& ViT-B/16 ($\Bc$1) &0.4644 & 0.4667 & 0.5066 & 0.5782 & 0.5865 & 0.5111\\
& ViT-B/16 ($\Bc$2) &0.5583 & 0.5652 & 0.5937  & 0.6802 & 0.6820 & 0.5954\\
& ViT-B/16 ($\Bc$3) &0.5744 & 0.5860 & 0.6103 & 0.6975 & 0.7081 & 0.6109\\
& ViT-B/16 ($\Bc$4) &0.5892 & 0.5908 & 0.6167 & 0.7015 & 0.7197 & 0.6231\\
& ViT-B/16 ($\Bc$5) &0.5909 & 0.5960 & 0.6250  & 0.7147 & 0.7201 & 0.6279\\
& ViT-B/16 ($\Bc$6) &0.6030 & 0.6034 & 0.6295 & 0.7196 & 0.7284 & 0.6350\\
\midrule
\multirow{3}{*}{TieredImageNet} 
& EfficientNet-B0 ($\Bc$4) & 0.6007 & 0.6048 & 0.6430 & 0.7620 & 0.7704 & 0.6210\\
& EfficientNet-B0 ($\Bc$5) & 0.6367 & 0.6359 & 0.6771 & 0.8123 & 0.8222 & 0.6904\\
& EfficientNet-B0 ($\Bc$6) & 0.6957 & 0.7096 & 0.7420 & 0.8270 & 0.8342 & 0.7249\\
% &\todo{effnet b1-b5, b2-b5} \\
% &\todo{Resnet101-b3} \\
\bottomrule
\end{tabular}
% }
\end{table}

%% file: sections/app-F-ablation.tex
In this section, we extend our ablation studies by showing the performance of \systemName when using 3 upstream model architectures and asymmetrical upstream model design. 

\subsection{Three ensembles}
\autoref{fig:ensemble_3_models} shows the scenario where we build an ensemble based on three upstream models. In this case, we require four downstream models for different model pairs, and in a situation where all three succeeded. \autoref{tab:app_3_ensembles} lists the accuracy of \systemName when training three upstream models across different architectures on CIFAR-100. The results show that: \emph{Adding more models always increased the overall accuracy with little to no effects on the individual upstream models or the downstream models}. For example, using three upstream models for EfficientNet-B0 ($\Bc$5) increases the accuracy by 1.44\%, while typically being able to retain the accuracy of the upstream models, as in the case of two models.
\input{tables-app/app-3-ensembles}

\subsection{Asymmetrical upstream models}
Resource availability may be heterogeneous as highlighted in \autoref{fig:app_freg}, requiring the models to have different upstream sizes. To address this and as a primary step to the limitations introduced in \autoref{sec:conclusion}, we discuss \autoref{tab:app_uneven} \emph{that demonstrates \systemName flexibility to build asymmetric ensemble models for diverse and resource-fragmented scenarios}. The table shows the accuracy of multiple scenarios when training on the CIFAR-100 dataset. As shown, our training approach highlights that the trained models are not only able to refine each other, but this performance is also similar to a symmetrically designed ensemble model. As a specific example, EfficientNet-B0 ($\Bc$4-$\Bc$5) achieves within 1\% of the performance as compared to EfficientNet-B0 ($\Bc$5) from \autoref{tab:arxiv_generalization} while utilizing asymmetrically distributed, but similar number of parameters (memory budget).
\input{tables-app/app-uneven}

%% file: tables-app/app-3-ensembles.tex
\begin{table}[t]
\centering
\scriptsize
% \caption{Accuracy results across different data sets and architectures.}
\caption{Performance of three upstream based \systemName models. Top-1 Acc. is reported for classification task on CIFAR-100 on (1) individual upstream models (\emph{e.g.} $h_{\{1\}}$) (2) ``2-combination'' downstream model (i.e. when two of the models are available \emph{e.g.} $h_{\{1,2\}}$) and (3) the downstream model when all the upstreams are available $h_{\{1,2,3\}}$.} 
\label{tab:app_3_ensembles}
% \resizebox{\textwidth}{!}{%
\begin{tabular}{ccccr|cccr|cr}
\toprule
\multirow{3}{*}{DNN Model} &  \multicolumn{10}{c}{\systemName}\\
\cmidrule{2-11}
% & & \multicolumn{2}{c}{$h_{\{1\}}$} & \multicolumn{2}{c}{$h_{\{2\}}$}  & \multicolumn{2}{c}{$h_{\{1,2\}}$}  & &  \\ 
& $h_{\{1\}}$ & $h_{\{2\}}$ & $h_{\{3\}}$ & Para. & $h_{\{1,2\}}$ & $h_{\{1,3\}}$ & $h_{\{2,3\}}$ & Para. & $h_{\{1,2,3\}}$ & Para.\\
\midrule
EfficientNet-B0 ($\Bc$1) & 0.1965 & 0.1852 & 0.2149 & 4.08K & 0.2249 & 0.2527 & 0.2449 & 11.45K & 0.2675 & 27.03K\\
EfficientNet-B0 ($\Bc$2) & 0.4036 & 0.4181 & 0.4458 & 21.59K & 0.4615 & 0.4819 & 0.4816 & 48.08K & 0.5010 & 86.77K\\
EfficientNet-B0 ($\Bc$3) & 0.5692 & 0.5781 & 0.5835 & 0.07M & 0.6205 & 0.6281 & 0.6274 & 0.15M & 0.6481 & 0.24M\\
EfficientNet-B0 ($\Bc$4) & 0.6780 & 0.6859 & 0.6887 & 0.32M & 0.7242 & 0.7263 & 0.7284 & 0.65M & 0.7443 & 1.02M\\
EfficientNet-B0 ($\Bc$5) & 0.6938 & 0.7209 & 0.7477 & 0.88M & 0.7552 & 0.7751 & 0.7811 & 1.79M & \textbf{0.7913} & 2.69M\\
EfficientNet-B0 ($\Bc$6) & 0.7273 & 0.7578 & 0.7502 & 2.90M & 0.7870 & 0.7808 & 0.7953 & 5.83M & \textbf{0.8042} & 8.86M\\
Resnet50 ($\Bc$1)  & 0.4962 & 0.5680 & 0.5691 & 0.25M & 0.5760 & 0.5911 & 0.5955 & 0.55M & 0.6182 & 0.98M \\
Resnet50 ($\Bc$2)  & 0.6442 & 0.6328 & 0.6805 & 1.40M & 0.6712 & 0.7030 & 0.6987 & 3.30M & 0.7152 & 4.95M\\
Resnet50 ($\Bc$3)  & 0.7470 & 0.6552 & 0.6977 & 8.76M & 0.7654 & 0.7612 & 0.7280 & 17.50M & \textbf{0.7761} & 26.86M\\
ViT-B/16 ($\Bc$1)  & 0.4497 & 0.4446 & 0.4564 & 7.80M & 0.4965 & 0.4956 & 0.4975 & 15.90M & 0.5199 & 24.42M\\
ViT-B/16 ($\Bc$2)  & 0.5495 & 0.5596 & 0.5450 & 14.90M & 0.5882 & 0.5826 & 0.5855 & 30.00M & 0.6053 & 45.68M\\
ViT-B/16 ($\Bc$3)  & 0.5586 & 0.5674 & 0.5706 & 22.00M & 0.5959 & 0.5985 & 0.6039 & 44.30M & 0.6163 & 66.93M\\
ViT-B/16 ($\Bc$4)  & 0.5802 & 0.5840 & 0.5784 & 29.17M & 0.6132 & 0.6055 & 0.6089 & 58.49M & \textbf{0.6227} & 88.19M\\                        
\bottomrule
\end{tabular}
% }
\end{table}

%% file: tables-app/app-uneven.tex
\begin{table}[t]
\centering
\scriptsize
% \caption{Accuracy results across different data sets and architectures.}
\caption{Asymmetric \systemName evaluation on CIFAR-100. Performance of MEL with uneven sized upstream models on the CIFAR-100 dataset. $^\dagger$Top-1 accuracy is reported for the image classification task.} 
\label{tab:app_uneven}
% \resizebox{\textwidth}{!}{%
\begin{tabular}{ccrcrcr}
\toprule
\multirow{3}{*}{DNN Model} &  \multicolumn{6}{c}{\systemName} \\
\cmidrule{2-7}
& \multicolumn{2}{c}{$h_{\{1\}}$} & \multicolumn{2}{c}{$h_{\{2\}}$}  & \multicolumn{2}{c}{$h_{\{1,2\}}$} \\ 
& Perf.$^\dagger$ & Para. & Perf.$^\dagger$ & Para. & Perf.$^\dagger$ & Para.\\
\midrule
EfficientNet-B0 ($\Bc$3-$\Bc$4) & 0.5687 & 0.07M & 0.7064 & 0.32M & 0.7116 & 0.40M  \\
EfficientNet-B0 ($\Bc$3-$\Bc$5) & 0.5690 & 0.07M & 0.7558 & 0.88M & 0.7624 & 0.95M \\
EfficientNet-B0 ($\Bc$4-$\Bc$5) & 0.6723 & 0.32M & 0.7540 & 0.88M & 0.7704 & 1.20M  \\
\midrule
%\todo{Resnet 50 B1}
% Resnet50 ($\Bc$1-$\Bc$2) & 0.5404 & 0.25M & 0.6890 & 1.50M & 0.6880 & 1.82M  \\
Resnet50 ($\Bc$1-$\Bc$3) & 0.5453 & 0.25M & 0.7331 & 8.76M & 0.7359 & 9.02M  \\
Resnet50 ($\Bc$2-$\Bc$3) & 0.6404 & 1.40M & 0.7301 & 8.76M & 0.7382 & 10.30M  \\
\bottomrule
% \multicolumn{10}{l}{$^\dagger$Top-1 Acc. is reported for classification tasks (CIFAR-100, TieredImageNet, Speech Commands); Word Error Rate (WER) is reported}\\
% \multicolumn{10}{l}{for DeepSpeech2, and perplexity is reported for BookCorpus.}
\end{tabular}
% }
\end{table}

%% file: sections/app-G-baselines.tex
In this section, we expand the results of \autoref{sec:eval_baselines} to the case when the architecture consists of 3-upstream models.

\input{tables-app/app-3-ensemble-compressed}

\subsection{Single failover replicas}
\autoref{tab:app_3_ensemble_smaller} expands the results from \autoref{tab:arxiv_smaller_models} when considering three upstream models, showing the efficacy of our training approach. It highlights that our \emph{training approach not only creates resilient models, but also creates models that behave well and sometimes better than individually trained models.}

\input{tables-app/app-3-ensemble-strawman}

\subsection{Training strategies}
\autoref{tab:app_3_ensemble_strawman} shows the performance of \systemName across different training strategies with the same architecture for comparision in the case of ensemble models trained with three upstream models. We compare three training approaches: 1) Our proposed \systemName, 2) standalone model training on the final accuracy, and 3) individually training upstream and downstream models.  As shown, \emph{training the model using our \systemName approach creates a better model than training models as standalone or training them individually and then combining them}.

%% file: tables-app/app-3-ensemble-compressed.tex
\begin{table}[t]
\centering
\scriptsize
\caption{Performance of individual backup models $h_{\{1\}}$, $h_{\{2\}}$, $h_{\{3\}}$ under joint optimization within the MEL framework for three upstream models compared to individual training. }
\label{tab:app_3_ensemble_smaller}
\begin{tabular}{ccccc}
\toprule
\multirow{2}{*}{DNN Model} & \multicolumn{3}{c}{\systemName Perf.} & \multirow{2}{*}{Standalone} \\
\cmidrule{2-4}
 & $h_{\{1\}}$ & $h_{\{2\}}$ & $h_{\{3\}}$ &\\ 
\midrule 
EfficientNet-B0 ($\Bc$1) & 0.1965 & 0.1852 & \textbf{0.2149} & 0.2060 \\
EfficientNet-B0 ($\Bc$2) & 0.4036 & 0.4181 & \textbf{0.4458} & 0.4268 \\
 EfficientNet-B0 ($\Bc$3) & 0.5692 & 0.5781 & \textbf{0.5835} & 0.5809 \\
 EfficientNet-B0 ($\Bc$4) & 0.6780 & 0.6859 & 0.6887 & 0.7105 \\
 EfficientNet-B0 ($\Bc$5) & 0.6938 & 0.7209 & \textbf{0.7477} & 0.7471 \\
 EfficientNet-B0 ($\Bc$6) & 0.7273 & 0.7578 & 0.7502 & 0.7686 \\
\midrule
 Resnet50 ($\Bc$2) & 0.6442 & 0.6328 & \textbf{0.6805} & 0.6672  \\
 Resnet50 ($\Bc$3) & \textbf{0.7470} & 0.6552 & 0.6977 & 0.7300  \\
\midrule
 ViT-B/16 ($\Bc$1) & 0.4497 & 0.4446 & 0.4564 & 0.4803\\
 ViT-B/16 ($\Bc$2) & 0.5495 & 0.5596 & 0.5450 & 0.5732\\
 ViT-B/16 ($\Bc$3) & 0.5586 & 0.5674 & 0.5706 & 0.5931\\
 ViT-B/16 ($\Bc$4) & 0.5802 & 0.5840 & 0.5784 & 0.6044\\
\bottomrule
\end{tabular}
\end{table}

%% file: tables-app/app-3-ensemble-strawman.tex
\begin{table}[t]
\centering
\scriptsize
\caption{Three upstream \systemName. Performance of $h_{\{1,2,3\}}$ under different training strategies --{Standalone} trains only $h_{\{1,2,3\}}$, while Individually trained trains $h_{\{1\}}, h_{\{2\}}, h_{\{3\}}$ independently and then optimizes $h_{\{1,2,3\}}$ with frozen upstream models.}

\label{tab:app_3_ensemble_strawman}
\begin{tabular}{cccc}
\toprule
DNN Model & \systemName Perf. & Standalone & Ind. Trained \\
\midrule
EfficientNet-B0 ($\Bc$1) & 0.2675 & 0.2880 & 0.2244 \\
EfficientNet-B0 ($\Bc$2) & 0.5010 & 0.5038 & 0.4636 \\
EfficientNet-B0 ($\Bc$3) & \textbf{0.6481} & 0.6345 & 0.6202 \\
EfficientNet-B0 ($\Bc$4) & \textbf{0.7443} & 0.7120 & 0.7393 \\
EfficientNet-B0 ($\Bc$5) & \textbf{0.7913} & 0.7367 & 0.7809 \\
EfficientNet-B0 ($\Bc$6) & \textbf{0.8042} & 0.7545 & 0.8001 \\
\midrule
Resnet50 ($\Bc$2)& \textbf{0.7152} & 0.6225 & 0.6805 \\
Resnet50 ($\Bc$3)& \textbf{0.7761} & 0.6651 & 0.7515 \\
\midrule
ViT-B/16 ($\Bc$1)& \textbf{0.5199} & 0.5160 & 0.5121 \\
ViT-B/16 ($\Bc$2)& \textbf{0.6053} & 0.5868 & 0.6038 \\
ViT-B/16 ($\Bc$3)& 0.6163 & 0.6039 & 0.6196 \\
ViT-B/16 ($\Bc$4)& 0.6227 & 0.6041 & 0.6309 \\
\bottomrule
\end{tabular}
\end{table}

%% file: sections/app-H-systems.tex
\begin{figure}[t]
  \centering
  \hfill
  \begin{subfigure}[b]{0.32\textwidth}
    \includegraphics[width=\textwidth]{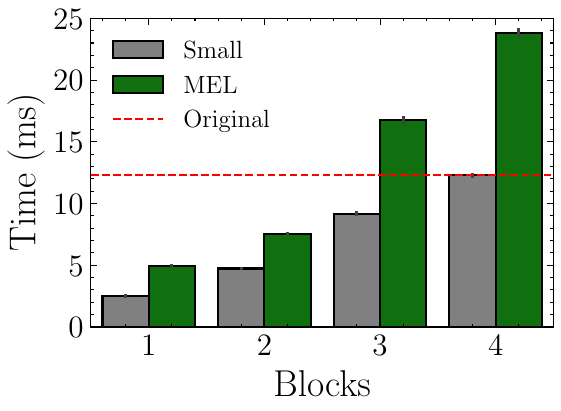}
    \caption{CPU}
    \label{fig:app_local_ResNet50_FC_CPU}
  \end{subfigure}
  \hfill
  \begin{subfigure}[b]{0.32\textwidth}
    \includegraphics[width=\textwidth]{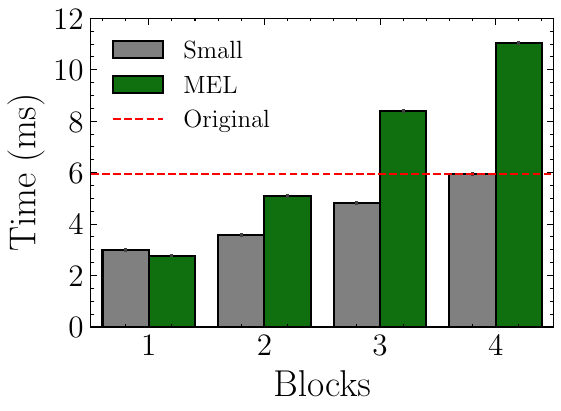}
    \caption{GPU}
    \label{fig:app_local_ResNet50_FC_GPU}
  \end{subfigure}
  \hfill
  \begin{subfigure}[b]{0.32\textwidth}
    \includegraphics[width=\textwidth]{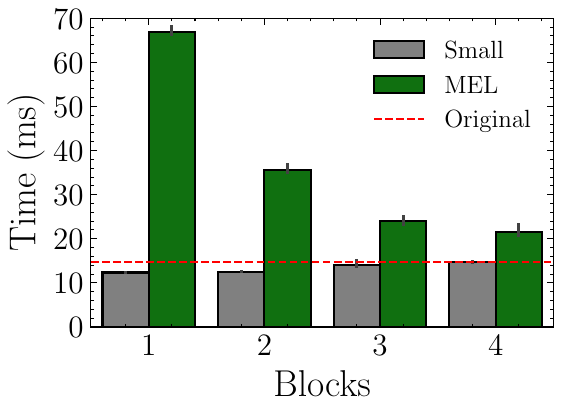}
    \caption{gRPC (GPU)}
    \label{fig:app_ERNet50_FC_rpc}
  \end{subfigure}
  \hfill
  \hfill
  \caption{Processing Latency of ResNet50 on CIFAR-100.}
  \label{fig:app_local_ResNet50_FC}
\end{figure}
\begin{figure}[t]
  \centering
  \hfill
  \begin{subfigure}[b]{0.32\textwidth}
    \includegraphics[width=\textwidth]{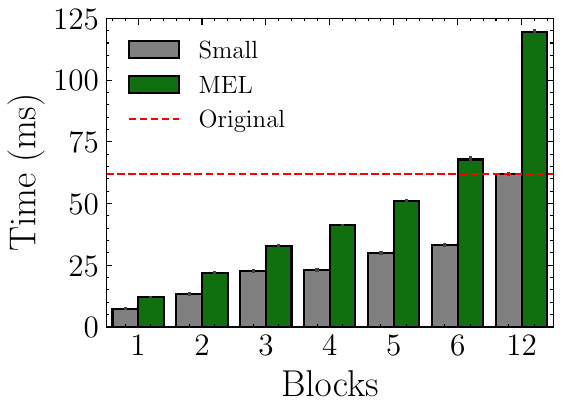}
    \caption{CPU}
    \label{fig:app_local_ViT_FC_CPU}
  \end{subfigure}
  \hfill
  \begin{subfigure}[b]{0.32\textwidth}
    \includegraphics[width=\textwidth]{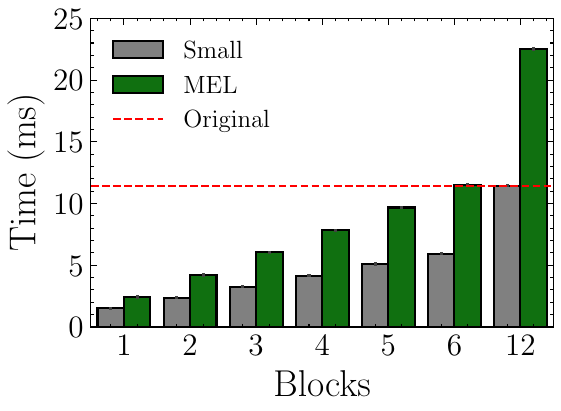}
    \caption{GPU}
    \label{fig:app_local_ViT_FC_GPU}
  \end{subfigure}
  \hfill
  \begin{subfigure}[b]{0.32\textwidth}
    \includegraphics[width=\textwidth]{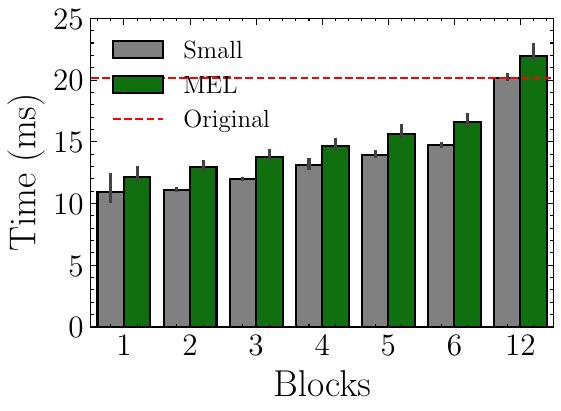}
    \caption{gRPC (GPU)}
    \label{fig:app_EViT_FC_rpc}
  \end{subfigure}
  \hfill
  \hfill
  \caption{Processing Latency of ViT-B/16 on CIFAR-100.}
  \label{fig:app_local_ViT_FC}
\end{figure}
\begin{figure}[t]
  \centering
  \hfill
  \begin{subfigure}[b]{0.32\textwidth}
    \includegraphics[width=\textwidth]{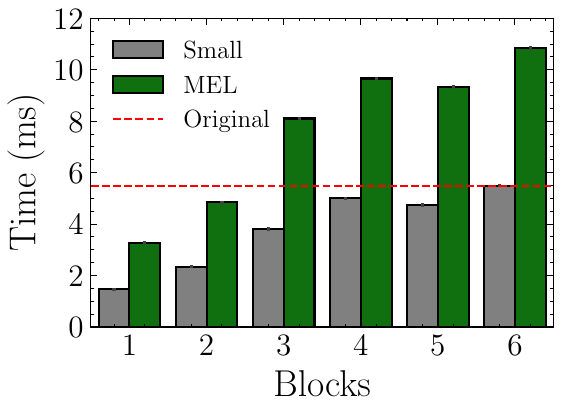}
    \caption{CPU}
    \label{fig:app_local_EDeepSp_FC_CPU}
  \end{subfigure}
  \hfill
  \begin{subfigure}[b]{0.32\textwidth}
    \includegraphics[width=\textwidth]{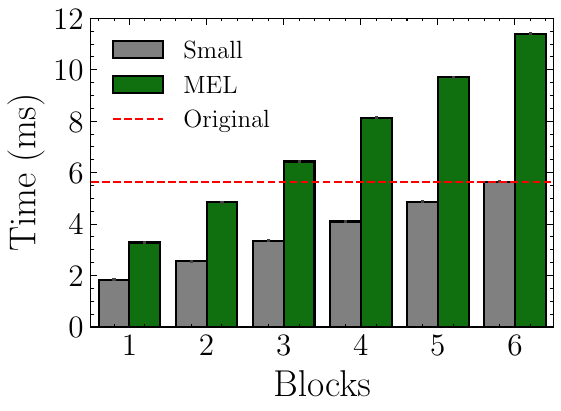}
    \caption{GPU}
    \label{fig:app_local_EDeepSp_FC_GPU}
  \end{subfigure}
  \hfill
  \begin{subfigure}[b]{0.32\textwidth}
    \includegraphics[width=\textwidth]{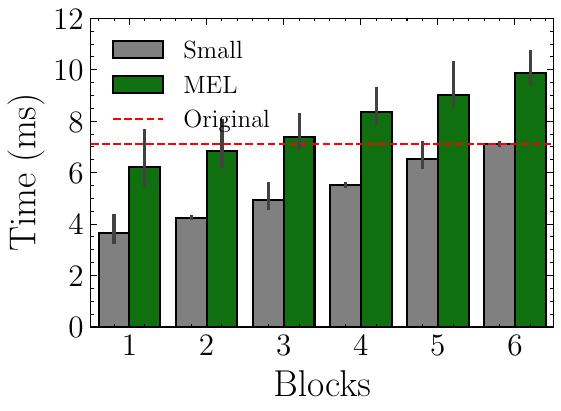}
    \caption{gRPC (GPU)}
    \label{fig:app_EDeepSp_FC_rpc}
  \end{subfigure}
  \hfill
  \hfill
  \caption{Processing Latency of DeepSpeech2 on Speech Commands.}
  \label{fig:app_local_EDeepSp_FC}
\end{figure}

In this section, we further evaluate the performance of our proposed multi-level ensemble models on CPU and the GPU, where ensemble models are hosted on the same server.
\autoref{fig:app_local_ResNet50_FC}, 
\autoref{fig:app_local_ViT_FC}, and 
\autoref{fig:app_local_EDeepSp_FC} shows the performance of the models when we deploy them across multiple servers. As shown, when deploying the models on one server, increasing the number of layers always increases the processing latency. For instance, going from one block to five blocks in EfficientNet-B0, see \autoref{fig:app_local_EENetB0_FC},  increases the latency by 4.8ms and 8.9ms for the small and the \systemName models, respectively.  
Moreover, the figures highlight that our proposed architecture typically uses more time than a small standalone model, as the ensemble models are much wider, hence requiring more processing time. 

Most importantly, the figures highlight the performance of the ensemble model when deployed on three servers (\emph{i.e.,} the intended deployment model). The results highlight a key aspect of deploying ensemble models. First, although using a smaller number of blocks leads to lower processing time and parameters. Our results, which are consistent with the results on split processing\cite{Huang2020:CLIO}, show that only using a small block prefix may lead to higher response time. Nonetheless, the results show that once we go past the first couple of layers, the results stabilize our models only include a small overhead.


% Time without RPC and response-time results across scenarios.